\documentclass[10pt,twocolumn,letterpaper]{article}

\usepackage{cvpr}
\usepackage{times}
\usepackage{epsfig}
\usepackage{graphicx}
\usepackage{amsmath}
\usepackage{amssymb}
\usepackage{bm}
\usepackage{algorithm}
\usepackage{algpseudocode}
\usepackage{subfigure}
\usepackage[toc,page]{appendix}
\usepackage[breaklinks=true,bookmarks=false]{hyperref}

\cvprfinalcopy 

\def\cvprPaperID{****} 

\begin{document}
\pagestyle{empty}
\title{Disentangling Latent Space for VAE by Label Relevant/Irrelevant Dimensions}

\author{{Zhilin Zheng$^{1}$ \quad Li Sun$^{1}$}\\
	{$^{1}$ Shanghai Key Laboratory of Multidimensional Information Processing, } \\ {East China Normal University} \\
	{\tt \small 51171214020@stu.ecnu.edu.cn \quad sunli@ee.ecnu.edu.cn}
}


\maketitle
\thispagestyle{empty}

\begin{abstract}
    VAE requires the standard Gaussian distribution as a prior in the latent space. Since all codes tend to follow the same prior, it often suffers the so-called "posterior collapse". To avoid this, this paper introduces the class specific distribution for the latent code. But different from cVAE, we present a method for disentangling the latent space into the label relevant and irrelevant dimensions, $\bm{\mathrm{z}}_s$ and $\bm{\mathrm{z}}_u$, for a single input. We apply two separated encoders to map the input into $\bm{\mathrm{z}}_s$ and $\bm{\mathrm{z}}_u$ respectively, and then give the concatenated code to the decoder to reconstruct the input. The label irrelevant code $\bm{\mathrm{z}}_u$ represent the common characteristics of all inputs, hence they are constrained by the standard Gaussian, and their encoder is trained in amortized variational inference way, like VAE. While $\bm{\mathrm{z}}_s$ is assumed to follow the Gaussian mixture distribution in which each component corresponds to a particular class. The parameters for the Gaussian components in $\bm{\mathrm{z}}_s$ encoder are optimized by the label supervision in a global stochastic way. In theory, we show that our method is actually equivalent to adding a KL divergence term on the joint distribution of $\bm{\mathrm{z}}_s$ and the class label $c$, and it can directly increase the mutual information between $\bm{\mathrm{z}}_s$ and the label $c$.
   Our model can also be extended to GAN by adding a discriminator in the pixel domain so that it produces high quality and diverse images.
\end{abstract}

\section{Introduction}

Learning a deep generative model for the structured image data is difficult because this task is not simply modeling a many-to-one mapping function such as the classification, instead it is often required to generate diverse outputs for similar codes sampled from a simple distribution. Furthermore, image $\bm{\mathrm{x}}$ in the high dimension space often lies in a complex manifold, thus the generative model should capture the underlying data distribution $p(\bm{\mathrm{x}})$.

Basically, Variational Auto-Encoder (VAE) \cite{rezende2014stochastic,kingma2014auto} and Generative Adversarial Network (GAN) \cite{goodfellow2014generative,makhzani2015adversarial} are two strategies for structured data generation. In VAE, the encoder $q_\phi(\bm{\mathrm{z}}|\bm{\mathrm{x}})$ maps data $\bm{\mathrm{x}}$ into the code $\bm{\mathrm{z}}$ in latent space. The decoder, represented by $p_\theta(\bm{\mathrm{x}}|\bm{\mathrm{z}})$, is given a latent code $\bm{\mathrm{z}}$ sampled from a distribution specified by the encoder and tries to reconstruct $\bm{\mathrm{x}}$. The encoder and decoder in VAE are trained together mainly based on the data reconstruction loss. At the same time, it requires to regularize the distribution $q_\phi(\bm{\mathrm{z}}|\bm{\mathrm{x}})$ to be simple (\emph{e.g.} Gaussian) based on the Kullback-Leibler (KL) divergence between $q(\bm{\mathrm{z}}|\bm{\mathrm{x}})$ and $p(\bm{\mathrm{z}})=\mathcal N(0,\bm{\mathrm{I}})$, so that the sampling in latent space is easy. Optimization for VAE is quite stable, but results from it are blurry. Mainly because the posterior defined by $q_\phi(\bm{\mathrm{z}}|\bm{\mathrm{x}})$ is not complex enough to capture the true posterior, also known for "posterior collapse". On the other hand, GAN treats the data generation task as a min/max game between a generator $G(\bm{\mathrm{z}})$ and discriminator $D(\bm{\mathrm{x}})$. The adversarial loss computed from the discriminator makes generated image more realistic, but its training becomes more unstable. In \cite{donahue2016adversarial,larsen2016autoencoding,mescheder2017adversarial}, VAE and GAN are integrated together so that they can benefit each other.

Both VAE and GAN work in an unsupervised way without giving any condition of the label on the generated image. Instead, conditional VAE (cVAE) \cite{sohn2015learning,bao2017cvae} extends it by showing the label $c$ for both encoder and decoder. It learns data distribution conditioned on the given label. Hence, the encoder and decoder become $q_\phi(\bm{\mathrm{z}}|\bm{\mathrm{x}},c)$ and $p_\theta(\bm{\mathrm{x}}|\bm{\mathrm{z}},c)$. Similarly, in conditional GAN (cGAN) \cite{chen2016infogan,isola2017image,odena2017conditional,miyato2018cgans} label $c$ is given to both generator $G(\bm{\mathrm{z}},c)$ and discriminator $D(\bm{\mathrm{x}},c)$. Theoretically, feeding label $c$ to either the encoder in VAE or decoder in VAE or GAN helps increasing the mutual information between the generated $\bm{\mathrm{x}}$ and the label $c$. Thus, it can improve the quality of generated image.

This paper deals with image generation problem in VAE with two separate encoders. For a single input $\bm{\mathrm{x}}$, our goal is to disentangle the latent space code $\bm{\mathrm{z}}$, computed by encoders, into the label relevant dimensions $\bm{\mathrm{z}}_s$ and irrelevant ones $\bm{\mathrm{z}}_u$. We emphasize the difference between $\bm{\mathrm{z}}_s$ and $\bm{\mathrm{z}}_u$, and their corresponding encoders. For $\bm{\mathrm{z}}_s$, since label $c$ is known during training, it should be more accurate and specific. While without any label constraint, $\bm{\mathrm{z}}_u$ should be general. Specifically, the two encoders are constrained with different priors on their posterior distributions $q_{\phi_s}(\bm{\mathrm{z}}_s|\bm{\mathrm{x}})$ and $q_{\phi_u}(\bm{\mathrm{z}}_u|\bm{\mathrm{x}})$. Similar with VAE or cVAE, in which the full code $\bm{\mathrm{z}}$ is label irrelevant, the prior for $\bm{\mathrm{z}}_u$ is also chosen $\mathcal N(0,\bm{\mathrm{I}})$. But different from previous works, the prior $p(\bm{\mathrm{z}}_s)$ becomes complex to capture the label relevant distribution. From the decoder's perspective, it takes the concatenation of $\bm{\mathrm{z}}_s$ and $\bm{\mathrm{z}}_u$ to reconstruct the input $\bm{\mathrm{x}}$. Here the distinction with cVAE and cGAN is that they uses the fixed, one-hot encoding label, while our work applies $\bm{\mathrm{z}}_s$, which is considered to be a variational, soft label. 

Note that there are two stages for training our model. First, the encoder for $\bm{\mathrm{z}}_s$ gets trained for classification task under the supervision of label $c$. Here instead of the softmax cross entropy loss, Gaussian mixture cross entropy loss proposed in \cite{wan2018rethinking} is adopted since it accumulates the mean $\bm{\mathrm{\mu}}_c$ and variance $\bm{\mathrm{\sigma}}_c$ for samples with the same label $c$, and models it as the Gaussian $\mathcal N(\bm{\mathrm{\mu}}_c,\bm{\mathrm{\sigma}}_c)$, hence $\bm{\mathrm{z}}_s\sim \mathcal N(\bm{\mathrm{\mu}}_c,\bm{\mathrm{\sigma}}_c)$. The first stage specifies the label relevant distribution. 
In the second stage, the two encoders and the decoder are trained jointly in an end-to-end manner based on the reconstruction loss. Meanwhile, priors of $\bm{\mathrm{z}}_s\sim\mathcal N(\bm{\mathrm{\mu}}_c,\bm{\mathrm{\sigma}}_c)$ and $\bm{\mathrm{z}}_u\sim\mathcal N(0,\bm{\mathrm{I}})$ are also considered. 

The main contribution of this paper lies in following aspects: (1) for a single input $\bm{\mathrm{x}}$ to the encoder, we provide an algorithm to disentangle the latent space into label relevant and irrelevant dimensions in VAE. Previous works like \cite{hadad2018two,bao2018towards,shu2017neural} disentangle the latent space in AE not VAE. So it is impossible to make the inference from their model. Moreover, \cite{mathieu2016disentangling,bao2018towards,DRIT} requires at least two inputs for training. (2) we find the Gaussian mixture loss function is suitable way for estimating the parameters of the prior distribution, and it can be optimized in VAE framework. 
(3) we give both a theoretical derivation and a variety of detailed experiments to explain the effectiveness of our work.

\section{Related works}
Two types of methods for the structured image generation are VAE and GAN. VAE \cite{kingma2014auto} is a type of parametric model defined by $p_\theta(\bm{\mathrm{x}}|\bm{\mathrm{z}})$ and $q_\phi(\bm{\mathrm{z}}|\bm{\mathrm{x}})$, which employs the idea of variational inference to maximize the evidence lower bound (ELBO), as is shown in (\ref{eq:eq1}).
\begin{equation}\label{eq:eq1}
\log p(\bm{\mathrm{x}})\geq \mathbb{E}_{q_\phi(\bm{\mathrm{z}}|\bm{\mathrm{x}})}(\log p_\theta(\bm{\mathrm{x}}|\bm{\mathrm{z}}) )- D_\text{KL}(q_\phi(\bm{\mathrm{z}}|\bm{\mathrm{x}})||p(\bm{\mathrm{z}}))
\end{equation}
The right side of the above is the ELBO, which is the lower bound of maximum likelihood. In VAE, a differentiable encoder-decoder are connected, and they are parameterized by $\phi$ and $\theta$, respectively. $E_{q_\phi(\bm{\mathrm{z}}|\bm{\mathrm{x}})}(\log p_\theta(\bm{\mathrm{x}}|\bm{\mathrm{z}}))$ represents the end-to-end reconstruction loss, and $\text{KL}(q_\phi(\bm{\mathrm{z}}|\bm{\mathrm{x}})||p(\bm{\mathrm{z}}))$ is the KL divergence between the encoder's output distribution $q_\phi(\bm{\mathrm{z}}|\bm{\mathrm{x}})$ and the prior $p(\bm{\mathrm{z}})$, which is usually modeled by standard normal distribution $\mathcal N(0,\bm{\mathrm{I}})$. Note that VAE assumes that the posterior $q_\phi(\bm{\mathrm{z}}|\bm{\mathrm{x}})$ is of Gaussian, and the $\bm{\mathrm{\mu}}$ and $\bm{\mathrm{\sigma}}$ are estimated for every single input $\bm{\mathrm{x}}$ by the encoder. This strategy is named amortized variational inference (AVI), and it is more efficiency than stochastic variational inference (SVI) \cite{hoffman2013stochastic}.

VAE's advantage is that its loss is easy to optimize, but the simple prior in latent space may not capture the complex data patterns which often leads to the mode collapse in latent space. Moreover, VAE's code is hard to be interpreted. Thus, many works focus on improving VAE on these two aspects. cVAE \cite{sohn2015learning} adds the label vector as the input for both the encoder and decoder, so that the latent code and generated image are conditioned on the label, and potentially prevent the latent collapse. On the other hand, $\beta$-VAE \cite{Higgins2017beta,burgess2018understanding} is a unsupervised approach for the latent space disentanglement. It introduces a simple hyper-parameter $\beta$ to balance the two loss term in (\ref{eq:eq1}). A scheme named infinite mixture of VAEs is proposed and applied in semi-supervised generation \cite{abbasnejad2017infinite}. It uses multiple number of VAEs and combines them as a non-parametric mixture model. 
In \cite{kim2018semi}, the semi-amortized VAE is proposed. It combines AVI with SVI in VAE. Here the SVI estimates the distribution parameters on the whole training set, while the AVI in traditional VAE gives this estimation for a single input.

GAN \cite{goodfellow2014generative} is another technique to model the data distribution $p_D(\bm{\mathrm{x}})$. It starts from a random $\bm{\mathrm{z}}\sim p(\bm{\mathrm{z}})$, where $ p(\bm{\mathrm{z}})$ is simple, \emph{e.g.} Gaussian, and trains a transform network $g_\theta (\bm{\mathrm{z}})$ under the help of discriminator $D_\phi(\cdot)$ so that $p_{\theta}(\bm{\mathrm{z}})$ approximates $p_D(\bm{\mathrm{x}})$. The later works \cite{nowozin2016f,mao2017least,arjovsky2017wasserstein,gulrajani2017improved,miyato2018spectral} try to stabilize GAN's training. Traditional GAN works in a fully supervised manner, while cGAN \cite{isola2017image,odena2017conditional,miyato2018cgans,bousmalis2017unsupervised} aims to generate images conditioned on labels. In cGAN, the label is given as an input to both the generator and discriminator as a condition for the distribution. The encoder-decoder architecture like AE or VAE can also be used in GAN. In ALI \cite{dumoulin2016adversarially} and BiGAN \cite{donahue2016adversarial}, the encoder maps $\bm{\mathrm{x}}$ to $\bm{\mathrm{z}}$, while the decoder reverses it. The discriminator takes the pair of $\bm{\mathrm{z}}$ and $\bm{\mathrm{x}}$, and is trained to determine whether it comes from the encoder or decoder in an adversarial manner. In VAE-GAN \cite{larsen2016autoencoding,liu2017unsupervised}, VAE's generated data are improved by a discriminator. Similar idea also applies to cVAE in \cite{bao2017cvae}. VAE-GAN also applies in some specific applications like \cite{bao2018towards,ge2018fdgan}. 

Since code $\bm{\mathrm{z}}$ potentially affects the generated data, some works try to model its effect and disentangle the dimensions of $\bm{\mathrm{z}}$. InfoGAN \cite{chen2016infogan} reveals the effect of latent space code $c$ by maximizing the mutual information between $c$ and the synthetic data $g_\theta(\bm{\mathrm{z,c}})$. Its generator outputs $g_\theta(\bm{\mathrm{z}},c)$ which is inspected by the discriminator $D_\phi(\cdot)$. $D_\phi(\cdot)$ also tries to reconstruct the code $c$. In \cite{mathieu2016disentangling}, the latent dimension is disentangled in VAE based on the specified factors and unspecified ones, which is similar with our work. But its encoder takes multiple inputs, and the decoder combines codes from different inputs for reconstruction. The work in \cite{hadad2018two} modifies \cite{mathieu2016disentangling} by taking a single input. To stabilize training, its model is built in AE not VAE, hence it can't perform variational inference. Other works in \cite{shu2017neural,bao2018towards,DRIT} are also built in AE and more than two inputs. Moreover they only apply in a particular domain like face \cite{shu2017neural,bao2018towards} or image-to-image translation \cite{DRIT}, while our work is built in VAE and takes only a single input for a more general case.

\begin{figure}[t]
\begin{center}
\includegraphics[width=1.00\linewidth]{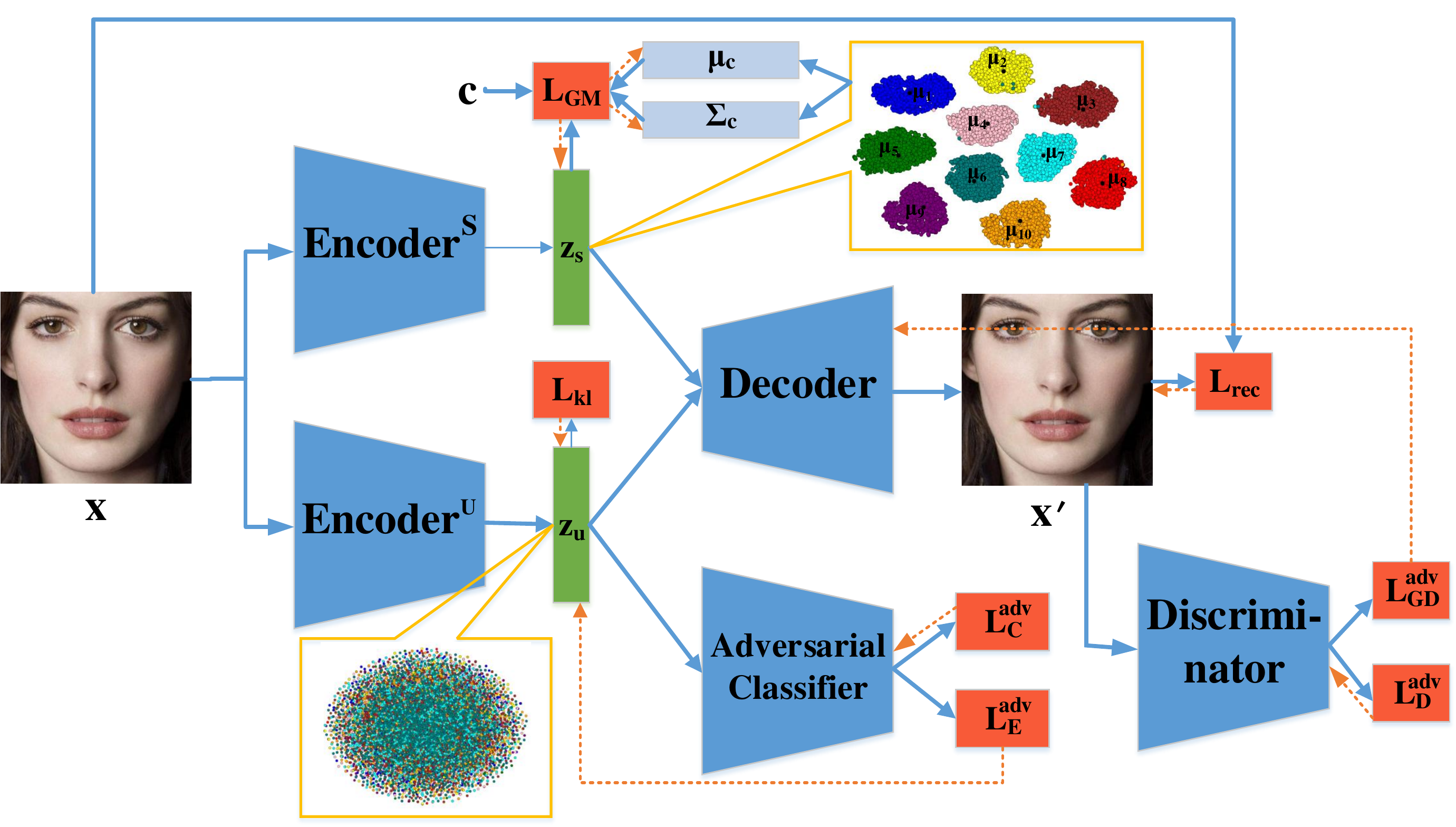}
\end{center}
   \caption{{\bf The network architecture}. We disentangle class relevant dimensions ${\bm{\mathrm{z}}_s}$ and class irrelevant dimensions ${\bm{\mathrm{z}}_u}$ in the latent space. The $Encoder^s$ maps input image ${\bm{\mathrm{x}}}$ to ${\bm{\mathrm{z}}_s}$, and forces ${\bm{\mathrm{z}}_s}$ to be well classified while following a Gaussian mixture distribution with learned mean ${\bm \mu_c}$ and covariance ${\bm \Sigma_c}$. Meanwhile, the $Encoder^u$ extracts ${\bm{\mathrm{z}}_u}$ from ${\bf x}$ and pushes it to match the standard Gaussian $\mathcal{N}({\bf 0},{\bf I})$. The adversarial classifier is added on the top of ${\bm{\mathrm{z}}_u}$ to distinguish the class of ${\bm{\mathrm{z}}_u}$, while $Encoder^u$ tries to fool it. Then ${\bm{\mathrm{z}}_s}$ and ${\bm{\mathrm{z}}_u}$ are concatenated and fed into the $Decoder$ to obtain ${\bf x'}$ for reconstruction. The adversarial training in the pixel domain is also adopted with a discriminator added on the images. The forward pass process is drawn in solid lines and dashed  lines represent back propagation.}
\label{fig:arch}
\end{figure}
\section{Proposed method}

We propose a image generation algorithm based on VAE which divides the encoder into two separate ones, one encoding label relevant representation $\bm{\mathrm{z}}_s$ and the other encoding label irrelevant information $\bm{\mathrm{z}}_u$. $\bm{\mathrm{z}}_s$ is learned with supervision of the categorical class label and it is required to follow a Gaussian mixture distribution, while $\bm{\mathrm{z}}_u$ is wished to contain other common information irrelevant to the label and is made close to standard Gaussian $\mathcal N({\bm 0},{\bm I})$.
\subsection{Problem formulation}
Given a labeled dataset $\mathcal{D}_s = \{({\bf x}^{1}, y^{1}), ({\bf x}^{2}, y^{2}),\cdots,({\bf x}^{(N)}, y^{(N)})\}$, where ${\bf x}^{(i)}$ is the $i$-th images and $y^{(i)} \in \{0,1,\cdots,C-1\}$ is the corresponding label. $C$ and $N$ are the number of classes and the size of the dataset, respectively. The goal of VAE is to maximum the ELBO defined in (\ref{eq:eq1}), so that the data log-likelihood $\log p(\bm{\mathrm{x}})$ is also maximized. The key idea is to split the full latent code ${\bf z}$ into the label relevant dimensions $\bm{\mathrm{z}}_s$ and the irrelevant dimensions ${\bf{z}_u}$, which means ${\bm{\mathrm{z}}_s}$ fully reflects the class $c$ but $\bm{\mathrm{z}}_u$ dose not. Thus the objective can be rewritten as (derived in detail in Appendices).
\begin{equation} \label{eq:eq2}
\begin{split}
\log p({\bf x}) &=\log \iint \sum_c p(\bm{\mathrm{x}}, \bm{\mathrm{z}}_s, \bm{\mathrm{z}}_u, c) d{\bm{\mathrm{z}}_s} d{\bm{\mathrm{z}}_u}\\
&\ge \mathbb{E}_{q_\psi(\bm{\mathrm{z}}_s|\bm{\mathrm{x}}),q_\phi(\bm{\mathrm{z}}_u|\bm{\mathrm{x}})}[\log p_\theta(\bm{\mathrm{x}} | \bm{\mathrm{z}}_s,\bm{\mathrm{z}}_u)] \\
&- D_\text{KL}(q_\phi(\bm{\mathrm{z}}_u|\bm{\mathrm{x}})||p(\bm{\mathrm{z}}_u))\\
&- D_\text{KL}(q_\psi(\bm{\mathrm{z}}_s, c|\bm{\mathrm{x}})||p(\bm{\mathrm{z}}_s, c))
\end{split}
\end{equation}

In Eq.~\ref{eq:eq2}, the ELBO becomes 3 terms in our setting. The \textbf{first} term is the negative reconstruction error, where $p_\theta$ is the decoder parameterized by $\theta$. It measures whether the latent code $\bm{\mathrm{z}}_s$ and $\bm{\mathrm{z}}_u$ are informative enough to recover the original data. 
In practice, the reconstruction error $L_{rec}$ can be defined as the $l_2$ loss between $\bf x$ and $\bf x'$. The \textbf{second} term acts as a regularization term of label irrelevant branch that pushes $q_\phi({\bm{\mathrm{z}}_u}|\bm{\mathrm{x}})$ to match the prior distribution $p(\bm{\mathrm{z}}_u)$, which is illustrated in detail in Section~\ref{sec:2.2}. The \textbf{third} term matches $q_\psi(\bm{\mathrm{z}}_s|\bm{\mathrm{x}})$ to a class-specific Gaussian distribution whose mean and covariance are learned with supervision, and it will be further introduced in Section \ref{sec:2.3}.

\subsection{Label irrelevant branch} \label{sec:2.2}

Intuitively, we want to disentangle the latent code $\bm{\mathrm{z}}$ into $\bm{\mathrm{z}}_s$ and $\bm{\mathrm{z}}_u$, and expect $\bm{\mathrm{z}}_u$ to follow a fixed, prior distribution which is irrelevant to the label. This regularization is realized by minimizing KL divergence between $q_\phi({\bm{\mathrm{z}}_u}|\bm{\mathrm{x}})$ and the prior $p(\bm{\mathrm{z}}_u)$ as illustrated in Eq.~\ref{eq:eq3}. More specifically, $q_\phi(\bm{\mathrm{z}}_u|\bm{\mathrm{x}})$ is a Gaussian distribution whose mean ${\bm\mu}$ and diagonal covariance ${\bm\Sigma}$ are the output of $Encoder^u$ parameterized by $\phi$. $p(\bm{\mathrm{z}}_u)$ is simply set to $N({\bm 0},{\bm I})$. Hence the KL regularization term is:
\begin{equation}\label{eq:eq3}
L_{kl} = D_\text{KL}[\mathcal{N}({\bm\mu},{\bm\Sigma})||\mathcal{N}({\bm 0},{\bm I})]
\end{equation}
Note that Eq.~\ref{eq:eq3} can be represented in a closed form, which is easy to be computed.

To ensure good disentanglement in $\bm{\mathrm{z}}_u$ and $\bm{\mathrm{z}}_s$, we introduce adversarial learning in the latent space as in AAE~\cite{makhzani2015adversarial} to drive the label relevant information out of $\bm{\mathrm{z}}_u$. To do this, an adversarial classifier is added on the top of $\bm{\mathrm{z}}_u$, which is trained to classify the category of $\bm{\mathrm{z}}_u$ with cross entropy loss as is shown in (\ref{eq:eq4}):
\begin{equation}\label{eq:eq4}
L_{C}^{adv} = - \mathbb{E}_{q_\phi(\bm{\mathrm{z}}_u|\bm{\mathrm{x}})}\sum_c \mathbb{I}(c=y) \log q_\omega(c | \bm{\mathrm{z}}_u)
\end{equation}
where $\mathbb{I}(c=y)$ is the indicator function, and $q_\omega(c | \bm{\mathrm{z}}_u)$ is softmax probability output by the adversarial classifier parameterized by $\omega$. Meanwhile, $Encoder^u$ is trained to fool the classifier, hence the target distribution becomes uniform over all categories, which is $\frac{1}{C}$. The cross entropy loss is defined as (\ref{eq:eq5}).
\begin{equation}\label{eq:eq5}
L_{E}^{adv} = - \mathbb{E}_{q_\phi(\bm{\mathrm{z}}_u|\bm{\mathrm{x}})}\sum_c \frac{1}{C} \log q_\omega(c | \bm{\mathrm{z}}_u)
\end{equation}

\subsection{Label relevant branch} \label{sec:2.3}

Inspired by GM loss~\cite{wan2018rethinking}, we expect $\bm{\mathrm{z}}_s$ to follow a Gaussian mixture distribution, expressed in Eq.~\ref{eq:eq6}, where ${\bm \mu_c}$ and ${\bm \Sigma_c}$ are the mean and covariance of Gaussian distribution for class $c$, and $p(c)$ is the prior probability, which is simply set to $\frac{1}{C}$ for all categories. For simplicity, we ignore the correlation among different dimensions of $\bm{\mathrm{z}}_s$, hence ${\bm \Sigma_c}$ is assumed to be diagonal.
\begin{equation} \label{eq:eq6}
p(\bm{\mathrm{z}}_s) = \sum_c p(\bm{\mathrm{z}}_s | c)p(c) = \sum_c \mathcal{N}(\bm{\mathrm{z}}_s;{\bm \mu_c},{\bm \Sigma_c})p(c)
\end{equation}
Recall that in Eq.~\ref{eq:eq2}, the KL divergence between $q_\psi(\bm{\mathrm{z}}_s, c|\bm{\mathrm{x}})$ and $p(\bm{\mathrm{z}}_s, c)$ is minimized. If ${\bm{\mathrm{z}}_s}$ is formulated as a Gaussian distribution with its ${\bm \Sigma} \to {\bf 0}$ and its mean $\hat{\bm{\mathrm{z}}}_s$ output by $Encoder^s$, which is actually a Dirac delta function $\delta(\bm{\mathrm{z}}_s-\hat{\bm{\mathrm{z}}}_s)$, the KL divergence turns out to be the likelihood regularization term $L_{lkd}$ in Eq.~\ref{eq:eq7}, which is proved in Appendices. Here $\bm \mu_y$ and $\bm \Sigma_y$ are the mean and covariance specified by the label $y$.
\begin{equation}\label{eq:eq7}
\begin{split}
L_{lkd} =- \log \mathcal{N}(\hat{\bm{\mathrm{z}}}_s;{\bm \mu_y},{\bm \Sigma_y})
\end{split}
\end{equation}

Furthermore, we want $\bm{\mathrm{z}}_s$ to contain label information as much as possible, thus the mutual information between $\bm{\mathrm{z}}_s$ and class $c$ is added to the maximization objective function. We prove in Appendices that it's equal to minimize the cross-entropy loss of the posterior probability $q(c | \bm{\mathrm{z}}_s)$ and the label, which is exactly the classification loss $L_{cls}$ in GM loss as is shown in  Eq.~\ref{eq:eq8}.
\begin{equation}\label{eq:eq8}
\begin{split}
L_{cls} &= -\mathbb{E}_{q_\psi(\bm{\mathrm{z}}_s|\bm{\mathrm{x}})} \sum_c \mathbb{I}(c=y)\log q(c | \bm{\mathrm{z}}_s) \\
&= -\log \frac{\mathcal{N}(\hat{\bm{\mathrm{z}}}_s | {\bm \mu_y},{\bm \Sigma_y})p(y)}{\sum_k \mathcal{N}(\hat{\bm{\mathrm{z}}}_s | {\bm \mu_k},{\bm \Sigma_k})p(k)}
\end{split}
\end{equation}
These two terms are added up to form GM loss in Eq.~\ref{eq:eq9}. Here $L_{GM}$ is finally used to train the $Encoder^s$.
\begin{equation}\label{eq:eq9}
L_{GM} = L_{cls} + \lambda_{lkd} L_{lkd}
\end{equation}
\subsection{The decoder and the adversarial discriminator}
The latent codes $\bm{\mathrm{z}}_s$ and $\bm{\mathrm{z}}_u$ output by $Encoder^s$ and $Encoder^u$ are first concatenated together, and then further given to the decoder to reconstruct the input $\bm{\mathrm{x}}$ by $\bm{\mathrm{x}}'$. Here the $Decoder$ is indicated by $p_\theta(\bm{\mathrm{x}}|\bm{\mathrm{z}})$ with its parameter $\theta$ learned from the $l_2$ reconstruction error $L_{rec}$. To synthesize a high quality $\bm{\mathrm{x}}'$, we also employ the adversarial training in the pixel domain. Specifically, a discriminator $D_{\theta_d}(\bm{\mathrm{x}},c)$ with adversarial training on its parameter $\theta_d$ is used to improve $\bm{\mathrm{x}}'$. Here the label $c$ is utilized in $D_{\theta_d}$ like in \cite{miyato2018cgans}. The adversarial training loss for discriminator can be formulated as in Eq.~\ref{eq:eq10},
\begin{equation}\label{eq:eq10}
\begin{split}
L_{D}^{adv} =& - \mathbb{E}_{\bm{\mathrm{x}}\sim P_r}[\log D_{\theta_d}(\bm{\mathrm{x}},c)]\\
&-\mathbb{E}_{\bm{\mathrm{z}}_u\sim \mathcal{ N}({\bm 0},{\bm I}),\bm{\mathrm{z}}_s\sim p(\bm{\mathrm{z}}_s)}[\log(1-D_{\theta_d}(G(\bm{\mathrm{z}}_s,\bm{\mathrm{z}}_u),c))]
\end{split}
\end{equation}
while this loss becomes $$L_{GD}^{adv}=-\mathbb{E}_{\bm{\mathrm{z}}_u\sim \mathcal{N}({\bm 0},{\bm I}),\bm{\mathrm{z}}_s\sim p(\bm{\mathrm{z}}_s)}[\log(D_{\theta_d}(G(\bm{\mathrm{z}}_s,\bm{\mathrm{z}}_u),c))]$$ for the generator.
Note that here $G(\bm{\mathrm{z}}_s,\bm{\mathrm{z}}_u)$ is the decoder and $p(\bm{\mathrm{z}}_s)$ is defined in Eq.~\ref{eq:eq6}.

\subsection{Training algorithm}

The training detail is illustrated in Algorithm~\ref{alg:1}. The $Encoder^s$, modeled by $q_\psi$, extracts label relevant code $\bm{\mathrm{z}}_s$. $Encoder^s$ is trained with $L_{GM}$ and $L_{rec}$, encouraging $\bm{\mathrm{z}}_s$ to be label dependent and follow a learned Gaussian mixture distribution. Meanwhile, the $Encoder^u$ represented by $q_\phi$ is intended to extract class irrelevant code $\bm{\mathrm{z}}_u$. It's trained by $L_{kl}$, $L_E^{adv}$ and $L_{rec}$ to make ${\bm{\mathrm{z}}_u}$ irrelevant to the label and be close to $\mathcal{N}({\bf 0}, {\bf I})$. The adversarial classifier parameterized by $\omega$ is learned to classify ${\bm{\mathrm{z}}_u}$ using $L_C^{adv}$. Then the decoder $p_\theta$ generates reconstruction image using the combined feature of $\bm{\mathrm{z}}_s$ and ${\bm{\mathrm{z}}_u}$ with the loss $L_{rec}$.

In the training process, a 2-stage alternating training algorithm is adopted. First, $Encoder^s$ is updated using $L_{GM}$ to learn mean ${\bm \mu_c}$ and covariance ${\bm \Sigma_c}$ of the prior $p(\bm{\mathrm{z}}_s | c)$. Then, the two encoders and the decoder are trained jointly to reconstruct images while the distributions of $\bm{\mathrm{z}}_s$ and ${\bm{\mathrm{z}}_u}$ are considered.
\begin{algorithm}[htb]
\caption{ The training process of our proposed architecture.}
\label{alg:1}
\begin{algorithmic}[1]
\Require
$\psi,\phi,\theta,\omega,\theta_d$ initial parameters of $Encoder^s$, $Encoder^u$, $Decoder$, the adversarial classifier on $\bm{\mathrm{z}}_u$ and the discriminator on $\bm{\mathrm{x}}$;
${\bm \mu_c}$ and ${\bm \Sigma_c}$ initial mean and covariance for Gaussian distribution of $\bm{\mathrm{z}}_s$;
$n_{gm}$, the number of iterations of $L_{GM}$ per end-to-end iteration;
$\lambda_{rec}$ and $\lambda_{kl}$ the weight of $L_{rec}$ and $L_{kl}$; 
\While {not converged}
\For{$i=0$ to $n_{gm}$}
\State Sample \{\bm{\mathrm{x}}, y\} a batch from dataset.
\State $\hat{\bf z}_s \gets Encoder^s(\bm{\mathrm{x}})$.
\State $L_{GM}\gets - \log q(y | \hat{\bf z}_s) - \lambda_{lkd}  \log p(\hat{\bf z}_s | y)$
\State $\psi \xleftarrow{+} - \nabla_\psi L_{GM}$
\State ${\bm \mu_c} \xleftarrow{+} - \nabla_{\bm \mu_c} L_{GM}, c \in [0,C-1]$
\State ${\bm \Sigma_c} \xleftarrow{+} - \nabla_{\bm \Sigma_c} L_{GM}, c \in [0,C-1]$
\EndFor
\State Sample \{\bm{\mathrm{x}}, y\} a batch from dataset.
\State ${\bm \mu}, {\bm \Sigma} \gets Encoder^u(\bm{\mathrm{x}})$
\State $L_{kl} \gets D_{KL}[\mathcal N({\bm\mu},{\bm\Sigma})||\mathcal N({\bm 0},{\bm I})]$
\State Sample ${\bf \epsilon}\sim \mathcal N({\bf 0}, {\bf I})$
\State ${\bf z}_u \gets  {\bm \Sigma}^{\frac{1}{2}}{\bm \epsilon} + {\bm \mu}$
\State $L_{E}^{adv} \gets - \sum_c \frac{1}{C} \log q_\omega(c | {\bf z}_u)$
\State $L_{C}^{adv} \gets - \log q_\omega(y | {\bf z}_u)$
\State $\hat{\bf z}_s \gets Encoder^s(\bm{\mathrm{x}})$.
\State $L_{lkd} \gets - \log \mathcal N(\hat{\bf z}_s;{\bm \mu_y},{\bm \Sigma_y})$
\State $\bm{\mathrm{x}}_f' \gets Decoder(\hat{\bf z}_s, {\bf z}_u)$
\State $L_{rec} \gets \frac{1}{2}||\bm{\mathrm{x}} - \bm{\mathrm{x}}_f'||_2^2$
\State Sample ${\bf z}_s^p \sim p({\bf z}_s | y),\ {\bf z}_u^p \sim \mathcal N({\bf 0}, {\bf I})$
\State $\bm{\mathrm{x}}_p' \gets Decoder({\bf z}_s^p, {\bf z}_u^p)$
\State $L_{D}^{adv} \gets - \log D_{\theta_d}(\bm{\mathrm{x}},y) - \log(1-D_{\theta_d}(\bm{\mathrm{x}}_f',y))-\log(1-D_{\theta_d}(\bm{\mathrm{x}}_p',y))$
\State $L_{GD}^{adv} \gets - \log(D_{\theta_d}(\bm{\mathrm{x}}_f',y)) - \log(D_{\theta_d}(\bm{\mathrm{x}}_p',y))$
\State $\psi \xleftarrow{+} - \nabla_\psi (L_{rec} + \lambda_{lkd} L_{lkd}) $
\State $\phi \xleftarrow{+} - \nabla_\phi(L_{E}^{adv} + \lambda_{kl}L_{kl} + \lambda_{rec}L_{rec} )$
\State $\omega \xleftarrow{+} - \nabla_\omega L_{C}^{adv}$
\State $\theta \xleftarrow{+} - \nabla_\theta (L_{rec}+ L_{GD}^{adv})$
\State $\theta_d \xleftarrow{+} - \nabla_{\theta_d} L_{D}^{adv}$
\EndWhile
\end{algorithmic}
\end{algorithm}


\subsection{Application in semi-supervised generation}\label{sec:semi}
Given $L$ unlabeled extra data $\mathcal{D}_u=\{\bm{\mathrm{x}}^{(N+1)},\bm{\mathrm{x}}^{(N+2)},\cdots,\bm{\mathrm{x}}^{(N+L)}\}$, we now use our architecture for the semi-supervised generation, in which the labels $y^{(N+i)}$ of $\bm{\mathrm{x}}^{(N+i)}$ in $\mathcal{D}_u$ are not presented. Here we hold the assumption that $\mathcal{D}_u$ are in the same domain as the fully supervised $\mathcal{D}_s$, but $y^{(N+i)}$ can be satisfied $y^{(N+i)} \in \{0,1,\cdots,C-1\}$, or out of the predefined range. In other words, if the absent $y^{(N+i)}$ is in the predefined range, its $\bm{\mathrm{z}}_s$ follows the same Gaussian mixture distribution as in Eq.~\ref{eq:eq6}. Otherwise, $\bm{\mathrm{z}}_s$ should follow an ambiguous Gaussian distribution defined in Eq.~\ref{eq:eq11}. 
\begin{equation} \label{eq:eq11}
\begin{split}
{\bm \mu_t} &= \sum_c p(c){\bm \mu_c} \\
{\bm \sigma_t^2} &= \sum_c p(c){\bm \sigma_c^2} + \sum_c p(c)({\bm \mu_c})^2 - (\sum_c p(c){\bm \mu_c})^2
\end{split}
\end{equation}
More specifically, $\bm{\mathrm{z}}_s$ is expected to follow $\mathcal N({\bm \mu_t} , {\bm \Sigma_t} )$ where ${\bm \mu_t}$ and ${\bm \Sigma_t}$ are the total mean and covariance of all the class-specific Gaussian distributions $\mathcal N({\bm \mu_c} , {\bm \Sigma_c} )$ as illustrated in Eq.~\ref{eq:eq6}. Here, ${\bm \Sigma_t}$ is diagonal matrix with ${\bm \sigma_t^2}$ as its variance vector. ${\bm \sigma_c^2}$ is also the variance vector of ${\bm \Sigma_c}$. Hence, the likelihood regularization term becomes $L_{lkd} =- \log \mathcal{N}(\hat{\bm{\mathrm{z}}_s};{\bm \mu_t},{\bm \Sigma_t})$. The whole network is trained in a end-to-end manner using total losses. Note that in this setting, the label $y$ is not provided, so $L_{GM}$ , $L^{adv}_E$ and $L^{adv}_C$ are ignored in the training process.

\section{Experiments}
In this section, experiments are carried out to validate the effectiveness of the proposed method. A toy example is first designed to show that by disentangling the label relevant and irrelevant codes, our model has the ability of generating diverse data samples than cVAE-GAN~\cite{bao2017cvae}. We then compare the quality of generated images on real image datasets. The latent space is also analyzed. Finally, the experiments of semi-supervised generation and image inpainting show the flexibility of our model, hence it may have many potential applications.

\subsection{Toy examples} \label{sec:4.1}
This section demonstrates our method on a toy example, in which the real data distribution lies in 2D with one dimension ($x$ axis) being label relevant and the other ($y$ axis) being irrelevant. The distribution is assumed to be known. There are 3 types of data points indicated by green, red and blue, belonging to 3 classes. The 2D data points and their corresponding labels are given to our model for variational inference and the new sample generation.

\begin{figure*}[htb]
\begin{center}
\includegraphics[width=1.00\linewidth]{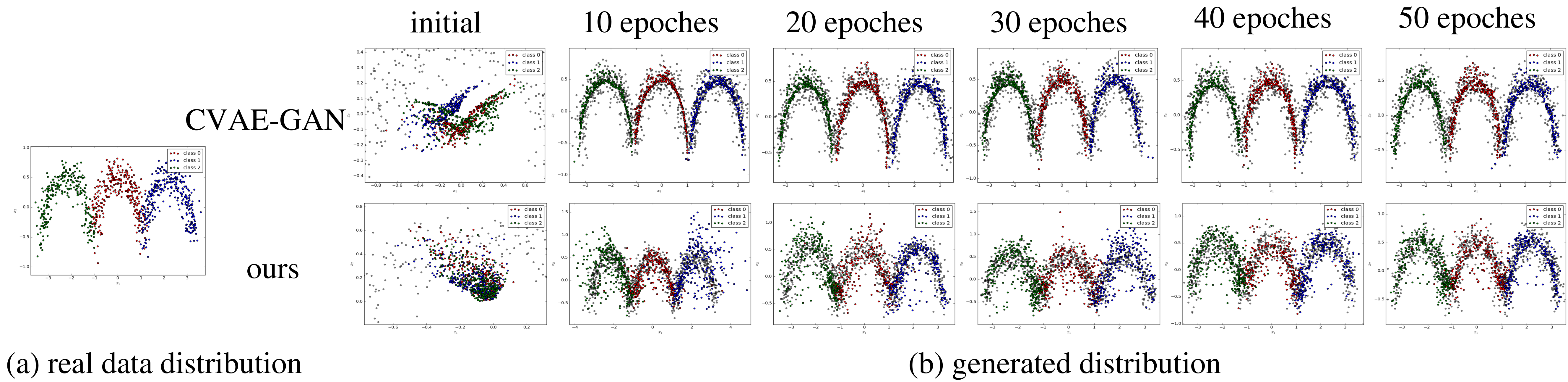}
\end{center}
   \caption{Results on a toy example for our model and cVAE-GAN. We show the generated points at different epochs.}
\label{fig:toyexample}
\end{figure*}

For comparison, we also give the same training data to cVAE-GAN for the same purpose. The two compared models share the similar settings of the network. In our model, the two encoders are both MLP with 3 hidden layers, and there are 32, 64, and 64 units in them. In cVAE-GAN, the encoder is the same, but it only has one encoder. The discriminators are exactly the same, which is also an MLP of 3 hidden layers with 32, 64, and 64 units. Adam is used as the optimization method in which a fixed learning rate of 0.0005 is applied for both. Each model is trained for 50 epochs until
they all converge. The generated samples of each model are plotted in Figure \ref{fig:toyexample}.

From Figure \ref{fig:toyexample} we can observe that both two models can capture the underlying data distribution, and our model converges at the similar rate. The advantage of our model is that it tends to generate diverse samples, while cVAE-GAN generates samples in a conserving way in which the label irrelevant dimensions are within the limited value range.

\begin{figure*}[htb]
\begin{center}
\includegraphics[width=1\linewidth]{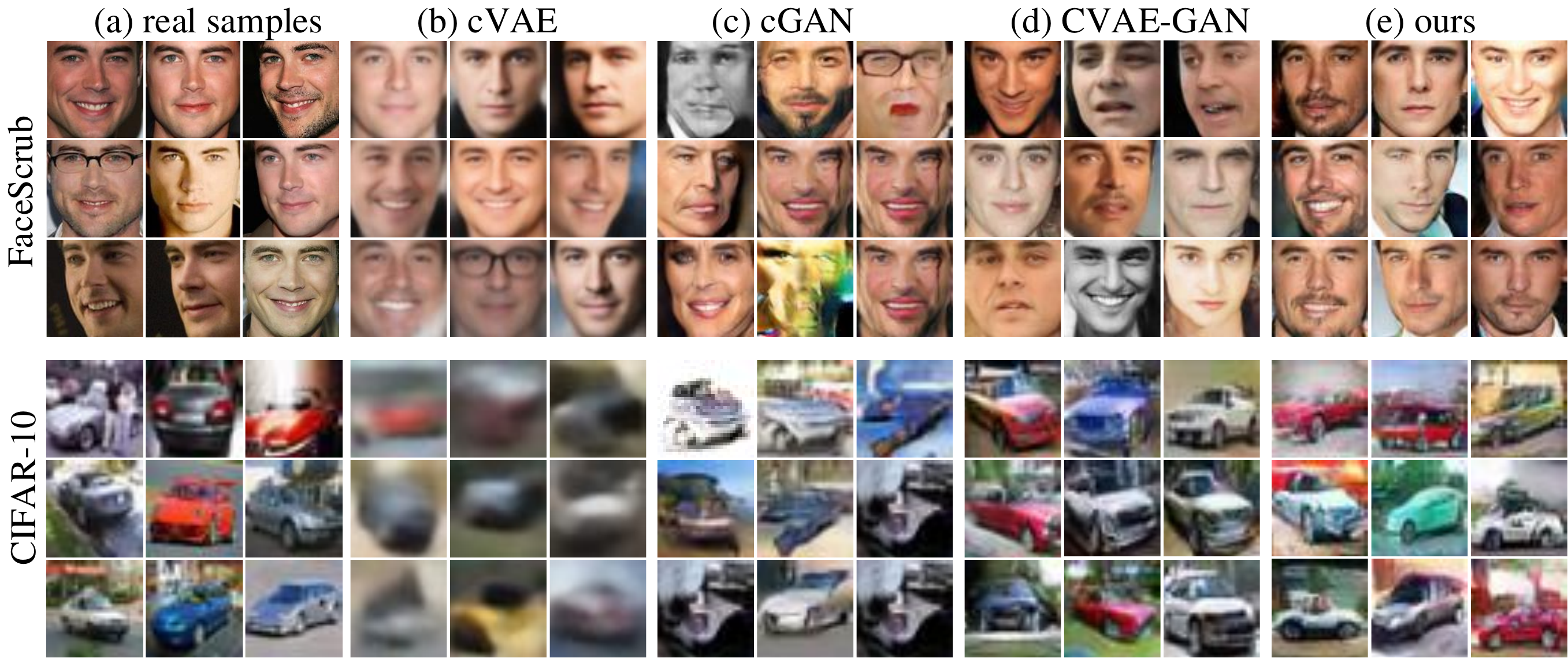}
\end{center}
   \caption{Visualization of generated images of different models.}
\label{fig:generated}
\end{figure*}

\subsection{Analysis on generated image quality} \label{sec:4.2}

In this section, we compare our method with other generative models for image generation quality. The experiments are conducted on two datasets: FaceScrub~\cite{ng2014data} and CIFAR-10~\cite{krizhevsky2009learning}. The FaceScrub contains $92k$ training images from $530$ different identities. For FaceScrub, a cascaded object detector proposed in~\cite{viola2001rapid} is first used to detect faces first, and then the face alignment is also conducted based on SDM proposed in~\cite{xiong2013supervised}. The detected cropped faces are resized to the fixed size 64$\times$64. In the training process, Adam optimizer with $\alpha=0.0005$ is used. The hyper parameter $\lambda_{lkd}$, $\lambda_{kl}$, and $\lambda_{rec}$ are set to 0.1, $\frac{10}{N_{pixel}}$, and $\frac{1}{N_{\bm{\mathrm{z}}_u}}$, respectively. Here, $N_{pixel}$ is the number of image pixels, and $N_{\bm{\mathrm{z}}_u}$ is the dimension of ${\bm{\mathrm{z}}_u}$. Since our method incorporates the label for training, popular generative networks conditioned on label, like cVAE~\cite{sohn2015learning}, cVAE-GAN~\cite{bao2017cvae}, and cGAN~\cite{miyato2018cgans}, are chosen for comparison. For cVAE, cVAE-GAN and cGAN, we randomly generate samples of class $c$ by first sampling ${\bf z} \sim \mathcal N({\bf 0}, {\bf I})$ and then concatenating ${\bf z}$ and one hot vector of $c$ as the input of decoder/generator. As for ours, $\bm{\mathrm{z}}_s\sim \mathcal N(\bm{\mathrm{\mu}}_c,\bm{\mathrm{\sigma}}_c)$ and $\bm{\mathrm{z}}_u\sim \mathcal N(\bm{\mathrm{0}},\bm{\mathrm{I}})$ are sampled and combined for decoder to generate samples. Some of generated images are visualized in Figure~\ref{fig:generated}. It shows that samples generated by cVAE are highly blurred, and cGAN suffers from mode collapse. Samples generated by cVAE-GAN and our method seem to have similar quality, we refer to two metrics, $Inception \ Score$~\cite{salimans2016improved} and intra-class diversity~\cite{ben2018gaussian} to compare them.

We adopt $Inception \ Score$ to evaluate realism and inter-class diversity of images. Generated images that are close to real images of class $y$ should have a posterior probability $p(y |\bm{\mathrm{x}})$ with low entropy. Meanwhile, images of diverse classes should have a marginal probability $p(y)$ with high entropy. Hence, $Inception \ Score$, formulated as $\exp(\mathbb{E}_{\bm{\mathrm{x}}}KL(p(y |\bm{\mathrm{x}})||p(y)))$, gets a high value when images are realistic and diverse.

To get conditional class probability $p(y |\bm{\mathrm{x}})$, we first train a classifier with Inception-ResNet-v1~\cite{szegedy2017inception} architecture on real data. Then we randomly generate 53k samples(100 for each class) of FaceScrub and 5k samples (500 for each class) of CIFAR-10, and apply them to the pre-trained classifier. The marginal $p(y)$ is obtained by averaging all $p(y |\bm{\mathrm{x}})$. The results are listed in Table~\ref{tab:1}.
\begin{table}
\begin{center}
\begin{tabular}{|l|c|c|}
\hline
 &FaceScrub& CIFAR-10\\
\hline\hline
 cVAE \cite{sohn2015learning}& 9.55 &  3.01\\
 cGAN~\cite{miyato2018cgans}&10.02 & 6.27\\
 cVAE-GAN~\cite{bao2017cvae}& 16.75 & 6.99\\
 ours& \bf{17.91} & \bf{7.04}\\
\hline
\end{tabular}
\end{center}
\caption{$Inception \ Score$ of different methods on two datasets. Please refer to~\ref{sec:4.2} for more details.}
\label{tab:1}
\end{table}

We emphasize that our method will generate more diverse samples in one class. Since $Inception \ Score$ only measures inter-class diversity, intra-class diversity of samples should also be taken into account. We adopt the metric proposed in ~\cite{ben2018gaussian}, which measures the average negative MS-SSIM~\cite{wang2003multiscale} between all pairs in the generated image set $\bm{X}$. Table ~\ref{tab:ssim} shows the inter-class diversity of cVAE-GAN and our method on FaceScrub and CIFAR-10.
\begin{equation}\label{eq:eq12}
d_{intra}(\bm{\mathrm{X}}) = 1- \frac{1}{|\bm{\mathrm{X}}|^2} \sum_{(\bm{\mathrm{x'}},\bm{\mathrm{x}}) \in \bm{\mathrm{X}} \times \bm{\mathrm{X}}} MS-SSIM(\bm{\mathrm{x'}},\bm{\mathrm{x}})
\end{equation}
\begin{table}
\begin{center}
\begin{tabular}{|l|c|c|}
\hline
 &FaceScrub& CIFAR-10\\
\hline\hline
 cVAE-GAN~\cite{bao2017cvae}& 0.0141 &0.0136 \\
 ours& \bf{0.0157} & \bf{0.0149}\\
\hline
\end{tabular}
\end{center}
\caption{Intra-class diversity of different methods on two datasets. Please refer to~\ref{sec:4.2} for more details.}
\label{tab:ssim}
\end{table}

\subsection{Analysis on disentangled latent space}
We now evaluate our proposal on the disentangled latent space, which is represented by label relevant dimensions $\bm{\mathrm{z}}_s$ and irrelevant ones $\bm{\mathrm{z}}_u$. $\bm{\mathrm{z}}_s$ for class $c$ is supposed to capture the variation unique to training images within the label $c$, while $\bm{\mathrm{z}}_u$ should contain the variation in common characteristics for all classes. It's validated in the following ways: (1) fixing $\bm{\mathrm{z}}_u$ and varying $\bm{\mathrm{z}}_s$. In this setting, we directly sample a $\bm{\mathrm{z}}_u\sim \mathcal N(\bm{\mathrm{0}},\bm{\mathrm{I}})$, and keep it fixed. Then a set of $\bm{\mathrm{z}}_s$ for class $c$ is obtained by first getting a series of random codes sampled from $\mathcal N(\bm{\mathrm{0}},\bm{\mathrm{I}})$ and then mapping them to class $c$. In specific, we first sample $\bm{\mathrm{z}}_1 \sim \mathcal N(\bm{\mathrm{0}},\bm{\mathrm{I}})$ and $\bm{\mathrm{z}}_2 \sim \mathcal N(\bm{\mathrm{0}},\bm{\mathrm{I}})$. Then a set of random codes $\bm{\mathrm{z}}^{(i)}$ are obtained by linear interpolation, i.e., $\bm{\mathrm{z}}^{(i)} = \alpha\bm{\mathrm{z}}_1 + (1-\alpha)\bm{\mathrm{z}}_2, \alpha \in [0,1]$. We map each $\bm{\mathrm{z}}^{(i)}$ to class $c$ with $\bm{\mathrm{z}}_s^{(i)} = \bm{\mathrm{z}}^{(i)} \odot \bm{\sigma}_c + \bm{\mu}_c$. Finally each $\bm{\mathrm{z}}_s^{(i)}$ is concatenated with the fixed $\bm{\mathrm{z}}_u$ and given to the decoder to get a generated image. (2) fixing $\bm{\mathrm{z}}_s$ and varying $\bm{\mathrm{z}}_u$. Similar to (1), we first sample a $\bm{\mathrm{z}}_s \sim \mathcal N(\bm{{\mu}_c},\bm{\sigma}_c)$ from a learned distribution and keep it fixed. Then a set of label irrelevant $\bm{\mathrm{z}}_u$ are obtained by linearly interpolating between $\bm{\mathrm{z}}_1$ and $\bm{\mathrm{z}}_2$, where $\bm{\mathrm{z}}_1$ and $\bm{\mathrm{z}}_2$ are sampled from $\mathcal N(\bm{\mathrm{0}},\bm{\mathrm{I}})$.

\begin{figure}[htb]
\subfigure[Fixing $\bm{\mathrm{z}}_u$ and varying $\bm{\mathrm{z}}_s$.]{
\begin{minipage}{0.6\linewidth}
\begin{center}
\includegraphics[width=1.6\linewidth]{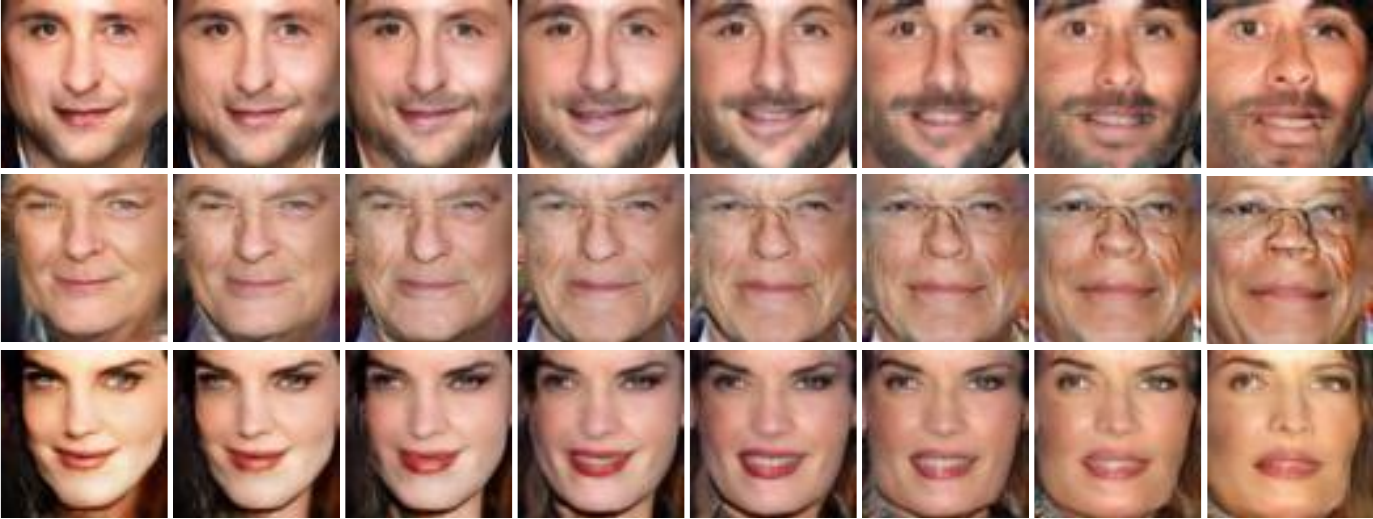}
\end{center}
\end{minipage}
}

\subfigure[Fixing $\bm{\mathrm{z}}_s$ and varying $\bm{\mathrm{z}}_u$.]{
\begin{minipage}{0.6\linewidth}
\begin{center}
\includegraphics[width=1.6\linewidth]{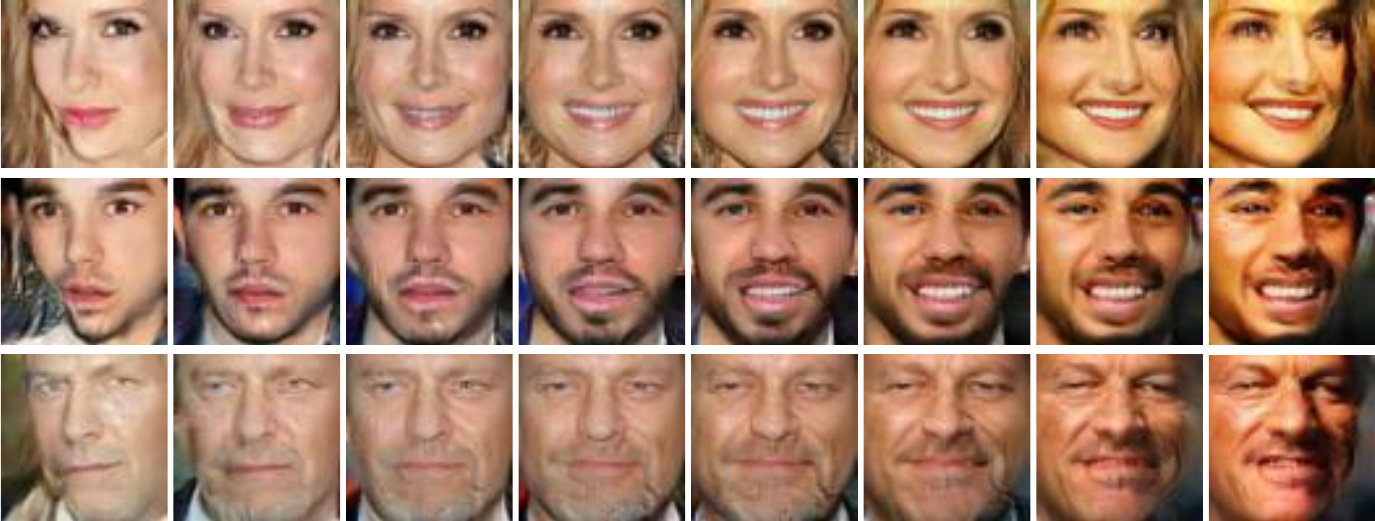}
\end{center}
\end{minipage}
}
 \caption{The generated images by fixing one code and varying the other. In (a), each row shows samples for linearly transformed $\bm{\mathrm{z}}_s$ of a certain class $c$ with a fixed $\bm{\mathrm{z}}_u$. In (b), each row corresponds to samples for linearly transformed $\bm{\mathrm{z}}_u$ with a fixed $\bm{\mathrm{z}}_s$ of class $c$.}
\label{fig:interpolation}
\end{figure}

We conduct experiments on FaceScrub and the generated images are shown in Figure~\ref{fig:interpolation}. In Figure~\ref{fig:interpolation} (a), each row presents samples generated by linearly transformed $\bm{\mathrm{z}}_s$ of a certain class $c$ and a fixed $\bm{\mathrm{z}}_u$. All three rows share the same $\bm{\mathrm{z}}_u$, and each column shares the same random code $\bm{\mathrm{z}}^{(i)}$ and just maps it to different class $c$ with $\bm{\mathrm{z}}_s^{(i)} = \bm{\mathrm{z}}^{(i)} \odot \bm{\sigma}_c + \bm{\mu}_c$. It shows that as $\bm{\mathrm{z}}_s$ varies, one may change differently for different identities, e.g., grow a beard, wrinkle, or take off the make-up. In Figure~\ref{fig:interpolation} (b), each row presents samples with linearly transformed $\bm{\mathrm{z}}_u$ a fixed $\bm{\mathrm{z}}_s$ of class $c$, and each column shares a same $\bm{\mathrm{z}}_u$. We can see that images from each row change consistently with poses, expressions and illuminations. These two experiments suggest that $\bm{\mathrm{z}}_s$ is relevant to $c$, while $\bm{\mathrm{z}}_u$ reflects more common label irrelevant characteristics.

We are also interested in each dimension in $\bm{\mathrm{z}}_u$ and conduct an experiment by varying a single element in it. We find three dimensions in $\bm{\mathrm{z}}_u$ which reflect the meaningful the common characteristics, such as the expression, elevation and azimuth.
\begin{figure*}[htb]
\begin{center}
\includegraphics[width=1.0\textwidth]{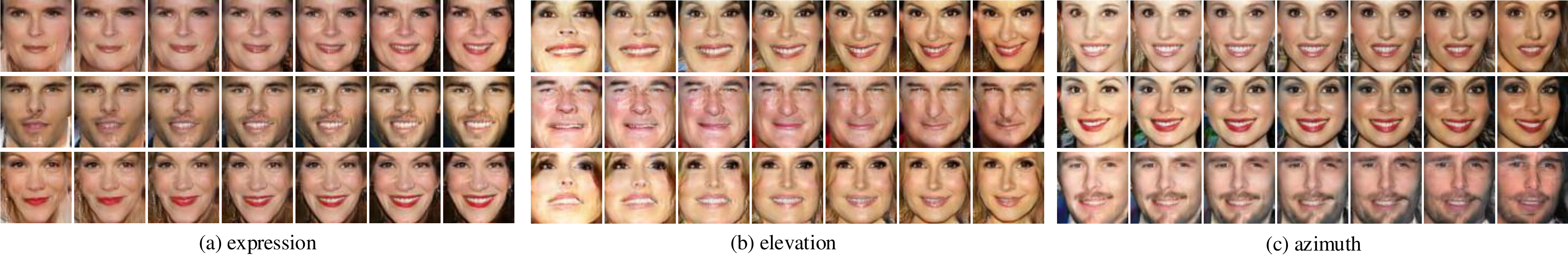}
\end{center}
   \caption{The generated images by fixing $\bm{\mathrm{z}}_s$ for each row and varying single dimensions in $\bm{\mathrm{z}}_u$. Here, we find three different dimensions in $\bm{\mathrm{z}}_u$, which directly causes the variations on expressions in (a), elevation in (b), and azimuth in (c).}
\label{fig:varyzuone}
\end{figure*}

\begin{figure*}[htb]
\begin{center}
\includegraphics[width=0.9\textwidth]{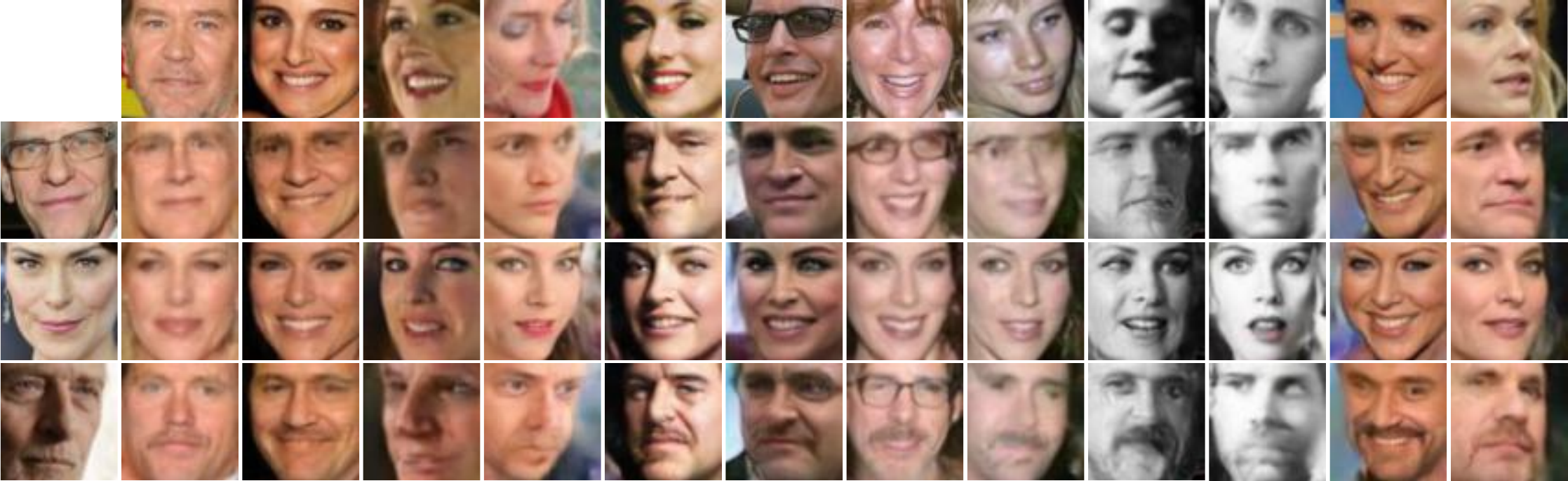}
\end{center}
   \caption{Face synthesis using images whose identities have not appeared in $\mathcal{D}_s$. Original images providing $\bm{\mathrm{z}}_u$ and $\bm{\mathrm{z}}_s$ are given in the first row and the first column. The synthesizing images using the combination of $\bm{\mathrm{z}}_u$ $\bm{\mathrm{z}}_s$ are shown in the corresponding position.}
\label{fig:synthesis}
\end{figure*}

\subsection{Semi-supervised image generation}
According to the details in Section~\ref{sec:semi}, the experiments on semi-supervised image generation are conducted. We find our method can learn well disentangled latent representation when the unlabeled extra data are available. To validate that, we randomly select 200 identities of about 21k images from CASIA~\cite{yi2014learning} dataset and remove their labels to form unlabeled dataset $\mathcal{D}_u$. Note that the identities in $\mathcal{D}_u$ are totally different with those in FaceScrub. After training the whole network on labeled dataset $\mathcal{D}_s$, we finetune it on $\mathcal{D}_u$ using the training algorithm illustrated in Section~\ref{sec:semi}.

To demonstrate the semi-supervised generation results, two different images are given to $Encoder^S$ and $Encoder^U$ to generate the code $\bm{\mathrm{z}}_s$ and $\bm{\mathrm{z}}_u$, respectively. Then, the decoder is required to synthesis a new image based on the concatenated code from $\bm{\mathrm{z}}_s$ and $\bm{\mathrm{z}}_u$. The Figure~\ref{fig:synthesis} shows face synthesis results using images whose identities have not appeared in $\mathcal{D}_s$. The first row and first column show a set of original images providing $\bm{\mathrm{z}}_u$ and $\bm{\mathrm{z}}_s$ respectively, while images in the middle are generated ones using $\bm{\mathrm{z}}_s$ of the corresponding row and $\bm{\mathrm{z}}_u$ of the corresponding column. It is obvious that the identity depends on $\bm{\mathrm{z}}_s$, while other characteristics like the poses, illumination, expressions are reflected on $\bm{\mathrm{z}}_u$. This semi-supervised generation shows $\bm{\mathrm{z}}_s$ and $\bm{\mathrm{z}}_u$ can also be disentangled on identities outside the labeled training data $\mathcal{D}_s$, which provides the great flexibility for image generation.

\subsection{Image inpainting}
\begin{figure}[htbp]
\begin{center}
\includegraphics[width=1.0\columnwidth]{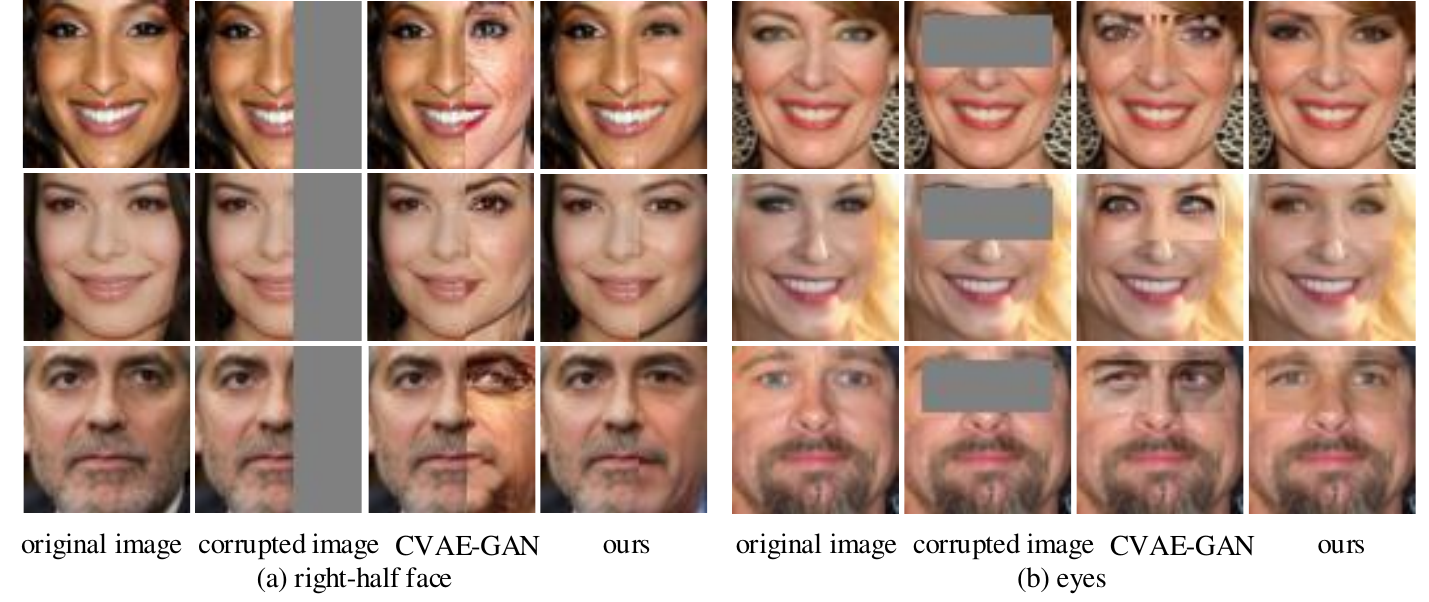}
\includegraphics[width=1.0\columnwidth]{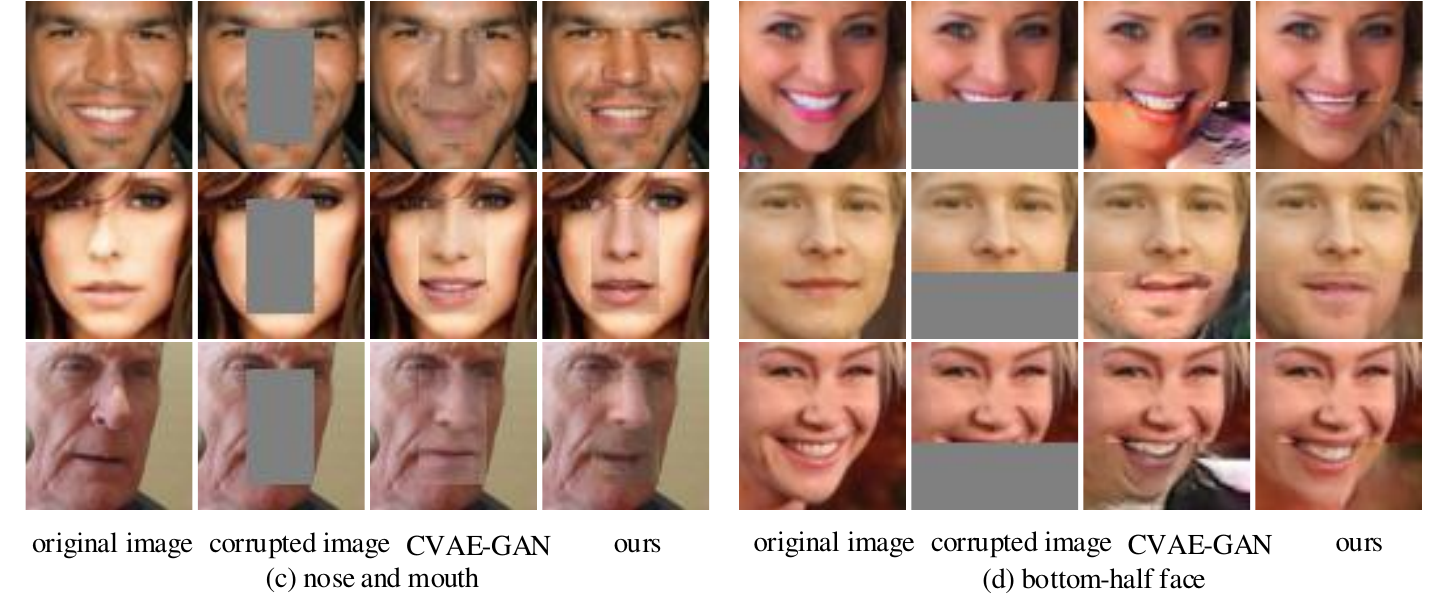}
\end{center}
   \caption{Image inpainting results. The original image is corrupted with different patterns, on the right-half, eyes, nose and mouth, and bottom-half face. We compare our model with cVAE-GAN.}
\label{fig:inpainting}\vspace{-0.5cm}
\end{figure}

Our method can also be applied to image inpainting. It means that given a partly corrupted image, we can extract meaningful latent code to reconstruct the original image. Note that in cVAE-GAN~\cite{bao2017cvae}, an extra class label $c$ should be provided for reconstruction while it's needless in our method. In practice, we first corrupt some patches for a image $\bm{x}$, namely right-half, eyes, nose and mouth, and bottom-half regions, then input those corrupted images into the two encoders to get $\bm{\mathrm{z}}_s$ and $\bm{\mathrm{z}}_u$, then the reconstructed image $\bm{x'}$ is generated using a combined $\bm{\mathrm{z}}_s$ and $\bm{\mathrm{z}}_u$. The image inpainting result is obtained by $\bm{x}^{inp} = \bm{M} \odot \bm{x'} + (1-\bm{M}) \odot \bm{x}$, where $\bm{M}$ is the binary mask for the corrupted patch. Figure~\ref{fig:inpainting} shows the results of image inpainting. cVAE-GAN struggles to complete the images when it comes to a large part of missing regions (e.g. right-half and bottom-half parts) or pivotal regions of faces (e.g. eyes), while our method provides visually pleasing results.

\section{Conclusion}
We propose a latent space disentangling algorithm on VAE baseline. Our model learns two separated encoders and divides the latent code into label relevant and irrelevant dimensions. Together with a discriminator in pixel domain, we show that our model can generate high quality and diverse images, and it can also be applied in semi-supervised image generation in which unlabeled data with unseen classes are given to the encoders. Future research includes building more interpretable latent dimensions with help of more labels, and reducing the correlation between the label relevant and irrelevant codes in our framework. 
\section*{Acknowledgements}
This work was supported in part by the National Natural Science Foundation of China under Project 61302125,  and in part by Natural Science Foundation of Shanghai under Project 17ZR1408500. Corresponding to sunli@ee.ecnu.edu.cn
{\small
\bibliographystyle{ieee}
\bibliography{egbib}

\begin{thebibliography}{10}\itemsep=-1pt

\bibitem{abbasnejad2017infinite}
M.~E. Abbasnejad, A.~Dick, and A.~van~den Hengel.
\newblock Infinite variational autoencoder for semi-supervised learning.
\newblock In {\em 2017 IEEE Conference on Computer Vision and Pattern
  Recognition (CVPR)}, pages 781--790. IEEE, 2017.

\bibitem{arjovsky2017wasserstein}
M.~Arjovsky, S.~Chintala, and L.~Bottou.
\newblock Wasserstein gan.
\newblock {\em stat}, 1050:9, 2017.

\bibitem{bao2017cvae}
J.~Bao, D.~Chen, F.~Wen, H.~Li, and G.~Hua.
\newblock Cvae-gan: Fine-grained image generation through asymmetric training.
\newblock In {\em 2017 IEEE International Conference on Computer Vision
  (ICCV)}, pages 2764--2773. IEEE, 2017.

\bibitem{bao2018towards}
J.~Bao, D.~Chen, F.~Wen, H.~Li, and G.~Hua.
\newblock Towards open-set identity preserving face synthesis.
\newblock In {\em Proceedings of the IEEE Conference on Computer Vision and
  Pattern Recognition}, pages 6713--6722, 2018.

\bibitem{ben2018gaussian}
M.~Ben-Yosef and D.~Weinshall.
\newblock Gaussian mixture generative adversarial networks for diverse
  datasets, and the unsupervised clustering of images.
\newblock {\em arXiv preprint arXiv:1808.10356}, 2018.

\bibitem{bousmalis2017unsupervised}
K.~Bousmalis, N.~Silberman, D.~Dohan, D.~Erhan, and D.~Krishnan.
\newblock Unsupervised pixel-level domain adaptation with generative
  adversarial networks.
\newblock In {\em The IEEE Conference on Computer Vision and Pattern
  Recognition (CVPR)}, volume~1, page~7, 2017.

\bibitem{burgess2018understanding}
C.~P. Burgess, I.~Higgins, A.~Pal, L.~Matthey, N.~Watters, G.~Desjardins, and
  A.~Lerchner.
\newblock Understanding disentangling in $\beta$-vae.
\newblock {\em arXiv preprint arXiv:1804.03599}, 2018.

\bibitem{cao2018vggface2}
Q.~Cao, L.~Shen, W.~Xie, O.~M. Parkhi, and A.~Zisserman.
\newblock Vggface2: A dataset for recognising faces across pose and age.
\newblock In {\em Automatic Face \& Gesture Recognition (FG 2018), 2018 13th
  IEEE International Conference on}, pages 67--74. IEEE, 2018.

\bibitem{chen2016infogan}
X.~Chen, Y.~Duan, R.~Houthooft, J.~Schulman, I.~Sutskever, and P.~Abbeel.
\newblock Infogan: Interpretable representation learning by information
  maximizing generative adversarial nets.
\newblock In {\em Advances in neural information processing systems}, pages
  2172--2180, 2016.

\bibitem{donahue2016adversarial}
J.~Donahue, P.~Kr{\"a}henb{\"u}hl, and T.~Darrell.
\newblock Adversarial feature learning.
\newblock {\em arXiv preprint arXiv:1605.09782}, 2016.

\bibitem{dumoulin2016adversarially}
V.~Dumoulin, I.~Belghazi, B.~Poole, O.~Mastropietro, A.~Lamb, M.~Arjovsky, and
  A.~Courville.
\newblock Adversarially learned inference.
\newblock In {\em ICLR}, 2017.

\bibitem{ge2018fdgan}
Y.~Ge, Z.~Li, H.~Zhao, G.~Yin, X.~Wang, and H.~Li.
\newblock Fd-gan: Pose-guided feature distilling gan for robust person
  re-identification.
\newblock In {\em Advances in Neural Information Processing Systems}, 2018.

\bibitem{goodfellow2014generative}
I.~Goodfellow, J.~Pouget-Abadie, M.~Mirza, B.~Xu, D.~Warde-Farley, S.~Ozair,
  A.~Courville, and Y.~Bengio.
\newblock Generative adversarial nets.
\newblock In {\em Advances in neural information processing systems}, pages
  2672--2680, 2014.

\bibitem{gulrajani2017improved}
I.~Gulrajani, F.~Ahmed, M.~Arjovsky, V.~Dumoulin, and A.~C. Courville.
\newblock Improved training of wasserstein gans.
\newblock In {\em Advances in Neural Information Processing Systems}, pages
  5767--5777, 2017.

\bibitem{hadad2018two}
N.~Hadad, L.~Wolf, and M.~Shahar.
\newblock A two-step disentanglement method.
\newblock In {\em Proceedings of the IEEE Conference on Computer Vision and
  Pattern Recognition}, pages 772--780, 2018.

\bibitem{Higgins2017beta}
I.~Higgins, L.~Matthey, A.~Pal, C.~Burgess, X.~Glorot, M.~Botvinick,
  S.~Mohamed, and A.~Lerchner.
\newblock $\beta$-vae: Learning basic visual concepts with a constrained
  variational framework.
\newblock In {\em ICLR}, 2017.

\bibitem{hoffman2013stochastic}
M.~D. Hoffman, D.~M. Blei, C.~Wang, and J.~Paisley.
\newblock Stochastic variational inference.
\newblock {\em The Journal of Machine Learning Research}, 14(1):1303--1347,
  2013.

\bibitem{isola2017image}
P.~Isola, J.-Y. Zhu, T.~Zhou, and A.~A. Efros.
\newblock Image-to-image translation with conditional adversarial networks.
\newblock {\em arXiv preprint}, 2017.

\bibitem{kim2018semi}
Y.~Kim, S.~Wiseman, A.~C. Miller, D.~Sontag, and A.~M. Rush.
\newblock Semi-amortized variational autoencoders.
\newblock {\em arXiv preprint arXiv:1802.02550}, 2018.

\bibitem{kingma2014auto}
D.~P. Kingma and M.~Welling.
\newblock Auto-encoding variational bayes.
\newblock {\em stat}, 1050:1, 2014.

\bibitem{krizhevsky2009learning}
A.~Krizhevsky and G.~Hinton.
\newblock Learning multiple layers of features from tiny images.
\newblock Technical report, Citeseer, 2009.

\bibitem{larsen2016autoencoding}
A.~B.~L. Larsen, S.~K. S{\o}nderby, H.~Larochelle, and O.~Winther.
\newblock Autoencoding beyond pixels using a learned similarity metric.
\newblock In {\em International Conference on Machine Learning}, pages
  1558--1566, 2016.

\bibitem{DRIT}
H.-Y. Lee, H.-Y. Tseng, J.-B. Huang, M.~K. Singh, and M.-H. Yang.
\newblock Diverse image-to-image translation via disentangled representations.
\newblock In {\em European Conference on Computer Vision}, 2018.

\bibitem{liu2017unsupervised}
M.-Y. Liu, T.~Breuel, and J.~Kautz.
\newblock Unsupervised image-to-image translation networks.
\newblock In {\em Advances in Neural Information Processing Systems}, pages
  700--708, 2017.

\bibitem{makhzani2015adversarial}
A.~Makhzani, J.~Shlens, N.~Jaitly, I.~Goodfellow, and B.~Frey.
\newblock Adversarial autoencoders.
\newblock {\em arXiv preprint arXiv:1511.05644}, 2015.

\bibitem{mao2017least}
X.~Mao, Q.~Li, H.~Xie, R.~Y. Lau, Z.~Wang, and S.~P. Smolley.
\newblock Least squares generative adversarial networks.
\newblock In {\em Computer Vision (ICCV), 2017 IEEE International Conference
  on}, pages 2813--2821. IEEE, 2017.

\bibitem{mathieu2016disentangling}
M.~F. Mathieu, J.~J. Zhao, J.~Zhao, A.~Ramesh, P.~Sprechmann, and Y.~LeCun.
\newblock Disentangling factors of variation in deep representation using
  adversarial training.
\newblock In {\em Advances in Neural Information Processing Systems}, pages
  5040--5048, 2016.

\bibitem{mescheder2017adversarial}
L.~Mescheder, S.~Nowozin, and A.~Geiger.
\newblock Adversarial variational bayes: Unifying variational autoencoders and
  generative adversarial networks.
\newblock In {\em International Conference on Machine Learning (ICML)}, pages
  2391--2400. PMLR, 2017.

\bibitem{miyato2018spectral}
T.~Miyato, T.~Kataoka, M.~Koyama, and Y.~Yoshida.
\newblock Spectral normalization for generative adversarial networks.
\newblock In {\em ICLR}, 2018.

\bibitem{miyato2018cgans}
T.~Miyato and M.~Koyama.
\newblock cgans with projection discriminator.
\newblock {\em arXiv preprint arXiv:1802.05637}, 2018.

\bibitem{ng2014data}
H.-W. Ng and S.~Winkler.
\newblock A data-driven approach to cleaning large face datasets.
\newblock In {\em Image Processing (ICIP), 2014 IEEE International Conference
  on}, pages 343--347. IEEE, 2014.

\bibitem{nowozin2016f}
S.~Nowozin, B.~Cseke, and R.~Tomioka.
\newblock f-gan: Training generative neural samplers using variational
  divergence minimization.
\newblock In {\em Advances in Neural Information Processing Systems}, pages
  271--279, 2016.

\bibitem{odena2017conditional}
A.~Odena, C.~Olah, and J.~Shlens.
\newblock Conditional image synthesis with auxiliary classifier gans.
\newblock In {\em International Conference on Machine Learning}, pages
  2642--2651, 2017.

\bibitem{rezende2014stochastic}
D.~J. Rezende, S.~Mohamed, and D.~Wierstra.
\newblock Stochastic backpropagation and approximate inference in deep
  generative models.
\newblock In {\em International Conference on Machine Learning}, pages
  1278--1286, 2014.

\bibitem{ILSVRC15}
O.~Russakovsky, J.~Deng, H.~Su, J.~Krause, S.~Satheesh, S.~Ma, Z.~Huang,
  A.~Karpathy, A.~Khosla, M.~Bernstein, A.~C. Berg, and L.~Fei-Fei.
\newblock {ImageNet Large Scale Visual Recognition Challenge}.
\newblock {\em International Journal of Computer Vision (IJCV)},
  115(3):211--252, 2015.

\bibitem{salimans2016improved}
T.~Salimans, I.~Goodfellow, W.~Zaremba, V.~Cheung, A.~Radford, and X.~Chen.
\newblock Improved techniques for training gans.
\newblock In {\em Advances in Neural Information Processing Systems}, pages
  2234--2242, 2016.

\bibitem{shu2017neural}
Z.~Shu, E.~Yumer, S.~Hadap, K.~Sunkavalli, E.~Shechtman, and D.~Samaras.
\newblock Neural face editing with intrinsic image disentangling.
\newblock In {\em Computer Vision and Pattern Recognition (CVPR), 2017 IEEE
  Conference on}, pages 5444--5453. IEEE, 2017.

\bibitem{simonyan2014very}
K.~Simonyan and A.~Zisserman.
\newblock Very deep convolutional networks for large-scale image recognition.
\newblock {\em arXiv preprint arXiv:1409.1556}, 2014.

\bibitem{sohn2015learning}
K.~Sohn, H.~Lee, and X.~Yan.
\newblock Learning structured output representation using deep conditional
  generative models.
\newblock In {\em Advances in Neural Information Processing Systems}, pages
  3483--3491, 2015.

\bibitem{szegedy2017inception}
C.~Szegedy, S.~Ioffe, V.~Vanhoucke, and A.~A. Alemi.
\newblock Inception-v4, inception-resnet and the impact of residual connections
  on learning.
\newblock In {\em AAAI}, volume~4, page~12, 2017.

\bibitem{szegedy2016rethinking}
C.~Szegedy, V.~Vanhoucke, S.~Ioffe, J.~Shlens, and Z.~Wojna.
\newblock Rethinking the inception architecture for computer vision.
\newblock In {\em Proceedings of the IEEE conference on computer vision and
  pattern recognition}, pages 2818--2826, 2016.

\bibitem{viola2001rapid}
P.~Viola and M.~Jones.
\newblock Rapid object detection using a boosted cascade of simple features.
\newblock In {\em Computer Vision and Pattern Recognition, 2001. CVPR 2001.
  Proceedings of the 2001 IEEE Computer Society Conference on}, volume~1, pages
  I--I. IEEE, 2001.

\bibitem{WahCUB_200_2011}
C.~Wah, S.~Branson, P.~Welinder, P.~Perona, and S.~Belongie.
\newblock {The Caltech-UCSD Birds-200-2011 Dataset}.
\newblock Technical Report CNS-TR-2011-001, California Institute of Technology,
  2011.

\bibitem{wan2018rethinking}
W.~Wan, Y.~Zhong, T.~Li, and J.~Chen.
\newblock Rethinking feature distribution for loss functions in image
  classification.
\newblock In {\em Proceedings of the IEEE Conference on Computer Vision and
  Pattern Recognition}, pages 9117--9126, 2018.

\bibitem{wang2003multiscale}
Z.~Wang, E.~P. Simoncelli, and A.~C. Bovik.
\newblock Multiscale structural similarity for image quality assessment.
\newblock In {\em The Thrity-Seventh Asilomar Conference on Signals, Systems \&
  Computers, 2003}, volume~2, pages 1398--1402. Ieee, 2003.

\bibitem{xiong2013supervised}
X.~Xiong and F.~De~la Torre.
\newblock Supervised descent method and its applications to face alignment.
\newblock In {\em Proceedings of the IEEE conference on computer vision and
  pattern recognition}, pages 532--539, 2013.

\bibitem{yi2014learning}
D.~Yi, Z.~Lei, S.~Liao, and S.~Z. Li.
\newblock Learning face representation from scratch.
\newblock {\em arXiv preprint arXiv:1411.7923}, 2014.

\end{thebibliography}
}

\clearpage

\begin{appendices}

\section{Mathematical proofs} 
\subsection{The ELBO of the log-likelihood objective}. 
We declare in Equation 2 that after dividing the latent space into label relevant dimensions $\bm{\mathrm{z}}_s$ and label irrelevant dimensions $\bm{\mathrm{z}}_u$, the ELBO of the log-likelihood objective $\log p(\bm{\mathrm{x}})$ becomes 3 terms in our setting. 
\begin{equation*}
\begin{split}
\log p(\bm{\mathrm{x}}) &=\log \iint p(\bm{\mathrm{x}}, \bm{\mathrm{z}}_s, \bm{\mathrm{z}}_u) d{\bm{\mathrm{z}}_s} d{\bm{\mathrm{z}}_u}\\
&\ge \mathbb{E}_{q_\psi(\bm{\mathrm{z}}_s|\bm{\mathrm{x}}),q_\phi(\bm{\mathrm{z}}_u|\bm{\mathrm{x}})}[\log p_\theta(\bm{\mathrm{x}} | \bm{\mathrm{z}}_s,\bm{\mathrm{z}}_u)] \\
&- D_\text{KL}(q_\phi(\bm{\mathrm{z}}_u|\bm{\mathrm{x}})||p(\bm{\mathrm{z}}_u))\\
&- D_\text{KL}(q_\psi(\bm{\mathrm{z}}_s, c|\bm{\mathrm{x}})||p(\bm{\mathrm{z}}_s, c))
\end{split}
\end{equation*}
{\bf Proof}

Our generative process is described as follows. First, sample a label relevant code $\bm{\mathrm{z}}_s \sim p(\bm{\mathrm{z}}_s , c)$ and a label irrelevant code $\bm{\mathrm{z}}_u \sim p(\bm{\mathrm{z}}_u)$. Then, a decoder $p_\theta(\bm{\mathrm{x}} | \bm{\mathrm{z}}_s,\bm{\mathrm{z}}_u)$, taking the combination of $\bm{\mathrm{z}}_s$ and $\bm{\mathrm{z}}_u$ as input, maps latent codes to images. Hence, we factorize the joint distribution $p(\bm{\mathrm{x}}, \bm{\mathrm{z}}_s, \bm{\mathrm{z}}_u)$ as:
\begin{equation*}
p(\bm{\mathrm{x}}, \bm{\mathrm{z}}_s, \bm{\mathrm{z}}_u) = \sum_c p_\theta(\bm{\mathrm{x}}| \bm{\mathrm{z}}_s, \bm{\mathrm{z}}_u) p(\bm{\mathrm{z}}_s , c) p(\bm{\mathrm{z}}_u)
\end{equation*}

By using Jensen’s inequality, the log-likelihood $\log p(\bm{\mathrm{x}})$ can
be written as:
\begin{equation*}
\begin{split}
\log p(\bm{\mathrm{x}}) &=\log \iint p(\bm{\mathrm{x}}, \bm{\mathrm{z}}_s, \bm{\mathrm{z}}_u) d{\bm{\mathrm{z}}_s} d{\bm{\mathrm{z}}_u}\\
&=\log \iint \sum_c p_\theta(\bm{\mathrm{x}}| \bm{\mathrm{z}}_s, \bm{\mathrm{z}}_u) p(\bm{\mathrm{z}}_s , c) p(\bm{\mathrm{z}}_u) d{\bm{\mathrm{z}}_s} d{\bm{\mathrm{z}}_u}\\
&= \log \mathbb{E}_{q_\psi(\bm{\mathrm{z}}_s, c|\bm{\mathrm{x}}), q_\phi(\bm{\mathrm{z}}_u|\bm{\mathrm{x}})} \frac{p_\theta(\bm{\mathrm{x}}| \bm{\mathrm{z}}_s, \bm{\mathrm{z}}_u)p(\bm{\mathrm{z}}_u)p(\bm{\mathrm{z}}_s, c)}{q_\psi(\bm{\mathrm{z}}_s, c|\bm{\mathrm{x}}) q_\phi(\bm{\mathrm{z}}_u|\bm{\mathrm{x}})}\\
&\ge \mathbb{E}_{q_\psi(\bm{\mathrm{z}}_s, c|\bm{\mathrm{x}}), q_\phi(\bm{\mathrm{z}}_u|\bm{\mathrm{x}})} [\log \frac{p_\theta(\bm{\mathrm{x}}| \bm{\mathrm{z}}_s,\bm{\mathrm{z}}_u) p(\bm{\mathrm{z}}_u) p(\bm{\mathrm{z}}_s, c)} {q_\psi(\bm{\mathrm{z}}_s, c|\bm{\mathrm{x}}) q_\phi(\bm{\mathrm{z}}_u|\bm{\mathrm{x}})}]\\
&= \mathbb{E}_{q_\psi(\bm{\mathrm{z}}_s, c|\bm{\mathrm{x}}), q_\phi(\bm{\mathrm{z}}_u|\bm{\mathrm{x}})} [\log p_\theta(\bm{\mathrm{x}}| \bm{\mathrm{z}}_s,\bm{\mathrm{z}}_u)]\\
&\ \ \ \ + \mathbb{E}_{q_\psi(\bm{\mathrm{z}}_s, c|\bm{\mathrm{x}}), q_\phi(\bm{\mathrm{z}}_u|\bm{\mathrm{x}})} [\log \frac{p(\bm{\mathrm{z}}_u)}{q_\phi(\bm{\mathrm{z}}_u|\bm{\mathrm{x}})}]\\
&\ \ \ \ + \mathbb{E}_{q_\psi(\bm{\mathrm{z}}_s, c|\bm{\mathrm{x}}), q_\phi(\bm{\mathrm{z}}_u|\bm{\mathrm{x}})} [\log \frac{p(\bm{\mathrm{z}}_s,c)}{q_\psi(\bm{\mathrm{z}}_s, c|\bm{\mathrm{x}})}]\\
&= \mathbb{E}_{q_\psi(\bm{\mathrm{z}}_s|\bm{\mathrm{x}}),q_\phi(\bm{\mathrm{z}}_u|\bm{\mathrm{x}})}[\log p_\theta(\bm{\mathrm{x}} | \bm{\mathrm{z}}_s,\bm{\mathrm{z}}_u)] \\
&\ \ \ \ - D_\text{KL}(q_\phi(\bm{\mathrm{z}}_u|\bm{\mathrm{x}})||p(\bm{\mathrm{z}}_u))\\
&\ \ \ \ - D_\text{KL}(q_\psi(\bm{\mathrm{z}}_s, c|\bm{\mathrm{x}})||p(\bm{\mathrm{z}}_s, c))
\end{split}
\end{equation*}

\subsection{Log-likelihood regularization term in the label relevant branch}
Note that the KL divergence $D_\text{KL}(q_\psi(\bm{\mathrm{z}}_s, c|\bm{\mathrm{x}})||p(\bm{\mathrm{z}}_s, c))$, the third term of the ELBO in Equation 2, is minimized. If we assume conditional independence between $\bm{\mathrm{z}}_s$ and the class $c$, then we have
\begin{equation*}
q_\psi(\bm{\mathrm{z}}_s, c|\bm{\mathrm{x}}) = q_\psi(\bm{\mathrm{z}}_s|\bm{\mathrm{x}}) p(c | x) 
\end{equation*}
where $p(c | x)$ is the one-hot encoding of the label $y$. If $q_\psi(\bm{\mathrm{z}}_s|\bm{\mathrm{x}})$ is formulated as Gaussian distribution with ${\bm \Sigma} \to {\bf 0}$ and mean $\hat{\bm{\mathrm{z}}}_s$ output by $Encoder^s$, which is actually a Dirac delta function.
\begin{equation*}
q_\psi(\bm{\mathrm{z}}_s|\bm{\mathrm{x}}) = \delta(\bm{\mathrm{z}}_s-\hat{\bm{\mathrm{z}}}_s)
\end{equation*}
The KL regularization term becomes
\begin{equation*}
\begin{split}
&D_\text{KL}(q_\psi(\bm{\mathrm{z}}_s, c|\bm{\mathrm{x}})||p(\bm{\mathrm{z}}_s, c))\\
=&D_\text{KL}[\delta(\bm{\mathrm{z}}_s - \hat{\bm{\mathrm{z}}}_s)p(c |\bm{\mathrm{x}})||p(\bm{\mathrm{z}}_s|c)p(c)] \\
=& -\sum_c \int \delta(\bm{\mathrm{z}}_s - \hat{\bm{\mathrm{z}}}_s)p(c |\bm{\mathrm{x}}) \log \frac{p(\bm{\mathrm{z}}_s|c)p(c)}{\delta(\bm{\mathrm{z}}_s -\hat{\bm{\mathrm{z}}}_s)p(c |{\bf x})} d\bm{\mathrm{z}}_s\\
=& -\sum_c\int \delta(\bm{\mathrm{z}}_s - \hat{\bm{\mathrm{z}}}_s)p(c |{\bf x}) \log p(\bm{\mathrm{z}}_s|c) d{\bm{\mathrm{z}}_s}\\
&- \sum_c\int \delta(\bm{\mathrm{z}}_s - \hat{\bm{\mathrm{z}}}_s)p(c |{\bf x}) \log \frac{p(c)}{p(c |{\bf x})}d\bm{\mathrm{z}}_s\\
&+ \sum_c\int \delta(\bm{\mathrm{z}}_s - \hat{\bm{\mathrm{z}}}_s)p(c |{\bf x}) \log \delta(\bm{\mathrm{z}}_s - \hat{\bm{\mathrm{z}}}_s)d\bm{\mathrm{z}}_s\\
=& -\sum_c p(c |{\bf x}) \log p(\hat{\bm{\mathrm{z}}}_s | c) -\sum_c p(c |{\bf x}) \log \frac{p(c)}{p(c |{\bf x})}\\
&+ \int \delta(\bm{\mathrm{z}}_s - \hat{\bm{\mathrm{z}}}_s) \log \delta(\bm{\mathrm{z}}_s - \hat{\bm{\mathrm{z}}}_s)d\bm{\mathrm{z}}_s
\end{split}
\end{equation*}

The second term relates to the prior distribution, so it can be regraded as a constant. The third term is negative entropy of delta function and has nothing to do with $\hat{\bm{\mathrm{z}}}_s$, hence we consider it as a constant too. Therefore, we have 
\begin{equation*}
\begin{split}
&D_\text{KL}(q_\psi(\bm{\mathrm{z}}_s, c|{\bf x})||p(\bm{\mathrm{z}}_s, c))\\
=& -\sum_c p(c |{\bf x}) \log p(\hat{\bm{\mathrm{z}}}_s | c) + Const. \\
=& -\sum_c \mathbb{I}(c=y) \log p(\hat{\bm{\mathrm{z}}}_s | c) + Const. 
\end{split}
\end{equation*}
where the prior distribution $p(\hat{\bm{\mathrm{z}}}_s | c)$ is set to $N(\hat{\bm{\mathrm{z}}}_s;{\bm \mu_c},{\bm \Sigma_c})$. Ignoring the constant term, it turns out to be the likelihood regularization term $L_{lkd}$ in Equation 7.
\begin{equation*}
\begin{split}
L_{lkd} &= - \sum_c \mathbb{I}(c=y) \log N(\hat{\bm{\mathrm{z}}}_s;{\bm \mu_c},{\bm \Sigma_c})\\
 &=- \log N(\hat{\bm{\mathrm{z}}}_s;{\bm \mu_y},{\bm \Sigma_y})
\end{split}
\end{equation*}
\subsection{Cross-entropy objective in the label relevant branch}
To encourage $\bm{\mathrm{z}}_s$ to become label relevant as much as possible, the mutual information $I(\bm{\mathrm{z}}_s;c)$ is maximized, where $\bm{\mathrm{z}}_s \sim q_\psi(\bm{\mathrm{z}}_s|{\bf x})$. In practice, $I(\bm{\mathrm{z}}_s;c)$ is hard to optimize directly because it requires access to $p(c | \bm{\mathrm{z}}_s)$. We can instead optimize its lower bound by introducing an auxiliary distribution $q(c | \bm{\mathrm{z}}_s)$ to approximate $p(c | \bm{\mathrm{z}}_s)$ as in infoGAN~\cite{chen2016infogan} .
\begin{equation*} 
\begin{split}
I(\bm{\mathrm{z}}_s;c) &= H(c) - H(c | \bm{\mathrm{z}}_s) \\
&= H(c) +  \mathbb{E}_{p({\bf x})}\mathbb{E}_{q_\psi(\bm{\mathrm{z}}_s|{\bf x})}\mathbb{E}_{p(c |\bm{\mathrm{z}}_s)} \log p(c |\bm{\mathrm{z}}_s) \\
&= H(c) + \mathbb{E}_{p({\bf x})} \mathbb{E}_{q_\psi(\bm{\mathrm{z}}_s|{\bf x})}\mathbb{E}_{p(c |\bm{\mathrm{z}}_s)} \log \frac{p(c |\bm{\mathrm{z}}_s)}{q(c |\bm{\mathrm{z}}_s)}\\
&+ \mathbb{E}_{p({\bf x})}\mathbb{E}_{q_\psi(\bm{\mathrm{z}}_s|{\bf x})}\mathbb{E}_{p(c |\bm{\mathrm{z}}_s)} \log q(c |\bm{\mathrm{z}}_s)\\
&\ge H(c) + \mathbb{E}_{p({\bf x})}\mathbb{E}_{q_\psi(\bm{\mathrm{z}}_s|{\bf x})}\mathbb{E}_{p(c |\bm{\mathrm{z}}_s)} \log q(c | \bm{\mathrm{z}}_s)
\end{split}
\end{equation*} 

Since we still need to sample from $p(c | \bm{\mathrm{z}}_s)$ in the inner expectation, we adopt Lemma 5.1 in infoGAN to further remove the need of $p(c | \bm{\mathrm{z}}_s)$. The first term of the lower bound is a constant, so we ignore it. Then the second term becomes
\begin{equation*} 
\begin{split}
& \mathbb{E}_{p({\bf x})}\mathbb{E}_{q_\psi(\bm{\mathrm{z}}_s|{\bf x})}\mathbb{E}_{p(c |\bm{\mathrm{z}}_s)} \log q(c | \bm{\mathrm{z}}_s) \\
=& \mathbb{E}_{p(c')}\mathbb{E}_{p({\bf x}|c')}\mathbb{E}_{q_\psi(\bm{\mathrm{z}}_s|{\bf x})}\mathbb{E}_{p(c |\bm{\mathrm{z}}_s)} \log q(c | \bm{\mathrm{z}}_s)  \\
=& \sum_{c'} \sum_c \iint p(c')p({\bf x}|c')q_\psi(\bm{\mathrm{z}}_s|{\bf x}) p(c |\bm{\mathrm{z}}_s) \log q(c | \bm{\mathrm{z}}_s) d{\bf x} d \bm{\mathrm{z}}_s \\
=& \sum_{c'} \sum_c \int p(c')p(c |\bm{\mathrm{z}}_s) \log q(c | \bm{\mathrm{z}}_s)[\int p({\bf x}|c')q_\psi(\bm{\mathrm{z}}_s|{\bf x}) d{\bf x}]d \bm{\mathrm{z}}_s 
\end{split}
\end{equation*} 

We hold the assumption that the process of sampling $\bm{\mathrm{z}}_s |{\bf x} $ is independent on $c$, thus 
\begin{equation*} 
\int p({\bf x}|c')q_\psi(\bm{\mathrm{z}}_s|{\bf x}) d{\bf x} = \int p(\bm{\mathrm{z}}_s, {\bf x}|c')d{\bf x} = p(\bm{\mathrm{z}}_s|c')
\end{equation*} 

According to Lemma 5.1 in infoGAN, we have
\begin{equation*} 
\begin{split}
&\sum_{c'} \sum_c \int p(c') p(\bm{\mathrm{z}}_s|c') p(c |\bm{\mathrm{z}}_s) \log q(c | \bm{\mathrm{z}}_s)d \bm{\mathrm{z}}_s\\
=& \sum_{c'} \int p(c') p(\bm{\mathrm{z}}_s|c') \log q(c' | \bm{\mathrm{z}}_s)d \bm{\mathrm{z}}_s
\end{split}
\end{equation*} 

Hence
\begin{equation*} 
\begin{split}
& \sum_{c'} \sum_c\int p(c')p(c |\bm{\mathrm{z}}_s) \log q(c | \bm{\mathrm{z}}_s)[\int p({\bf x}|c')q_\psi(\bm{\mathrm{z}}_s|{\bf x}) d{\bf x}]d \bm{\mathrm{z}}_s\\
=& \sum_{c'} \sum_c \int p(c') p(\bm{\mathrm{z}}_s|c') p(c |\bm{\mathrm{z}}_s) \log q(c | \bm{\mathrm{z}}_s)d \bm{\mathrm{z}}_s\\
=& \sum_c \int p(c) p(\bm{\mathrm{z}}_s|c) \log q(c | \bm{\mathrm{z}}_s)d \bm{\mathrm{z}}_s
\end{split}
\end{equation*} 

We further factorize $p(\bm{\mathrm{z}}_s|c)$ as $\int p({\bf x}|c)q_\psi(\bm{\mathrm{z}}_s|{\bf x}) d{\bf x}$, the equation above becomes
\begin{equation*} 
\begin{split}
&\sum_c \int p(c) p(\bm{\mathrm{z}}_s|c) \log q(c | \bm{\mathrm{z}}_s)d \bm{\mathrm{z}}_s \\
=& \sum_c \iint p(c) p({\bf x}|c)q_\psi(\bm{\mathrm{z}}_s|{\bf x}) \log q(c | \bm{\mathrm{z}}_s)d \bm{\mathrm{z}}_s d{\bf x} \\
=& \sum_c \iint p({\bf x})q_\psi(\bm{\mathrm{z}}_s|{\bf x}) p(c |{\bf x})\log q(c | \bm{\mathrm{z}}_s)d \bm{\mathrm{z}}_s d{\bf x} \\
=&\mathbb{E}_{p({\bf x})}\mathbb{E}_{q_\psi(\bm{\mathrm{z}}_s|{\bf x})} \sum_c p(c |{\bf x})\log q(c | \bm{\mathrm{z}}_s)
\end{split}
\end{equation*} 
where $p(c |{\bf x})$ is the one-hot encoding of the label $y$, i.e. $p(c |{\bf x}) = \mathbb{I}(c=y)$. To maximize it is to minimize its opposite, which is exactly the classification loss in Section 3.3.
\begin{equation*}
\begin{split}
L_{cls} = -\mathbb{E}_{p({\bf x})}\mathbb{E}_{q_\psi(\bm{\mathrm{z}}_s|{\bf x})} \sum_c \mathbb{I}(c=y)\log q(c | \bm{\mathrm{z}}_s)
\end{split}
\end{equation*} 


\section{Experimental details} 
\subsection{Dataset synthesis of toy example} 
Our synthetic dataset of toy example is a modification of the two-moon dataset, which contains three half circles instead of two. The generative process is described as follows. First, sample data points from three half unit circles with a horizontal interval of 2.2. Then, add Gaussian noises with $std=0.15$ to all of them.
\subsection{Network architecture of FaceScrub} 
For the two encoders, $Encoder^s$ and $Encoder^u$, we use VGG~\cite{simonyan2014very} architecture with batch normalization layers added to each layer and replace the last three fc layers with two fc layers of 1024 and 512 units. For the decoder, an inverse structure of the encoders is applied. The adversarial classifier in Section 3.2 consists of two fc layers of 256 and 530 units, and the discriminator contains 7 convolution layers and two fc layers (details are shown in Table~\ref{tab:1}). Note that  spectral normalization~\cite{miyato2018spectral} is applied to to the all of the weights in the discriminator and the label embedding is incorporated in the first fc layer as in~\cite{miyato2018cgans}.
\begin{table*}[htb]
\begin{center}
\begin{tabular}{|c|c|c|}
\hline
 Encoder & Decoder & Discriminator\\
\hline\hline
 input ${\bf x} \in \mathbb{R}^{32 \times 32 \times 3}$ & input $\bm{\mathrm{z}}_s \in \mathbb{R}^{100},\ \bm{\mathrm{z}}_u \in \mathbb{R}^{200}$ & input ${\bf x} \in \mathbb{R}^{32 \times 32 \times 3}$\\
 \hline
 $5 \times 5$ conv, 32, stride 2, batchnorm, relu & concat&$5 \times 5$ conv, 32, stride 1, lrelu\\
 \hline
 $5 \times 5$ conv, 64, stride 2, batchnorm, relu & fc, 1024, batchnorm, relu&$5 \times 5$ conv, 128, stride 2, lrelu\\
 \hline
 $3 \times 3$ conv, 128, stride 2, batchnorm, relu & $5 \times 5$ conv, 256, stride 2, batchnorm, relu&$5 \times 5$ conv, 256, stride 2, lrelu\\
 \hline
 $3 \times 3$ conv, 256, stride 2, batchnorm, relu & $5 \times 5$ conv, 256, stride 1, batchnorm, relu&$5 \times 5$ conv, 256, stride 2, lrelu\\
 \hline
 fc, 1024, batchnorm, relu  & $5 \times 5$ conv, 128, stride 2, batchnorm, relu& fc, 512, lrelu\\
 \hline
fc, 100 (for $\bm{\mathrm{z}}_s$) / 200 (for $\bm{\mathrm{z}}_u)$ & $5 \times 5$ conv, 64, stride 2, batchnorm, relu&fc, 1\\
 \hline
& $5 \times 5$ conv, 32, stride 2, batchnorm, relu&\\
 \hline
& $5 \times 5$ conv, 3, stride 1, tanh&\\
\hline
\end{tabular}
\end{center}
\caption{The network structure for Cifar-10.}
\label{tab:2}
\end{table*}
\begin{table}[htb]
\begin{center}
\begin{tabular}{|c|}
\hline
 Discriminator for FaceScrub\\
\hline\hline
 input ${\bf x} \in \mathbb{R}^{64 \times 64 \times 3}$\\
 \hline
 $3 \times 3$ conv, 64, stride 2, lrelu\\
 \hline
 $3 \times 3$ conv, 128, stride 2, lrelu\\
 \hline
 $3 \times 3$ conv, 256, stride 1, lrelu\\
 \hline
 $3 \times 3$ conv, 256, stride 2, lrelu\\
 \hline
 $3 \times 3$ conv, 512, stride 1, lrelu\\
 \hline
 $3 \times 3$ conv, 512, stride 2, lrelu\\
 \hline
 $3 \times 3$ conv, 512, stride 2, lrelu\\
 \hline
 global average pooling \\
 \hline
 fc, 1024, lrelu \\
 \hline
 fc, 1\\
\hline
\end{tabular}
\end{center}
\caption{The network structure of discriminator for FaceScrub.}
\label{tab:1}
\end{table}
\subsection{Network architecture of Cifar-10} 
The network structures of the two encoders, decoder and discriminator for Cifar-10 are shown in Table~\ref{tab:2}. The adversarial classifier in the latent space is similar as that used for FaceScrub, which are two fc layers of 256 and 10 units. Also, spectral normalization and label embedding are applied in the discriminator.

\subsection{Optimization}
We use Adam optimizer with $\alpha=0.0005$, $\beta_1=0$ and $\beta_2=0.9$. Since in the training process, the first stage using $L_{GM}$ is trained 3 times per second stage iteration, $L_{GM}$ converges fast. Continuously training after it converges will cause instability of $L_{GM}$ because ${\bm \Sigma_c}$ goes down gradually. In practice, we decay the learning rate of ${\bm \Sigma_c}$ by 0.01 after 2 epochs. 

\subsection{Inception Score}
Recall that $Inception\ Score$ requires access to the conditional class probability $p(y|{\bf x})$. We use classification model of Inception-ResNet-v1~\cite{szegedy2017inception} architecture trained on VGGFace2~\cite{cao2018vggface2} to evaluate generative models trained on FaceScrub. For generative models trained on Cifar-10, classification model of Inception-v3~\cite{szegedy2016rethinking} architecture trained on ImageNet~\cite{ILSVRC15} is used.

\section{Additional experiment results}
\subsection{More generated samples on FaceScrub and Cifar-10}
\begin{figure*}
\begin{center}
\includegraphics[width=1\linewidth]{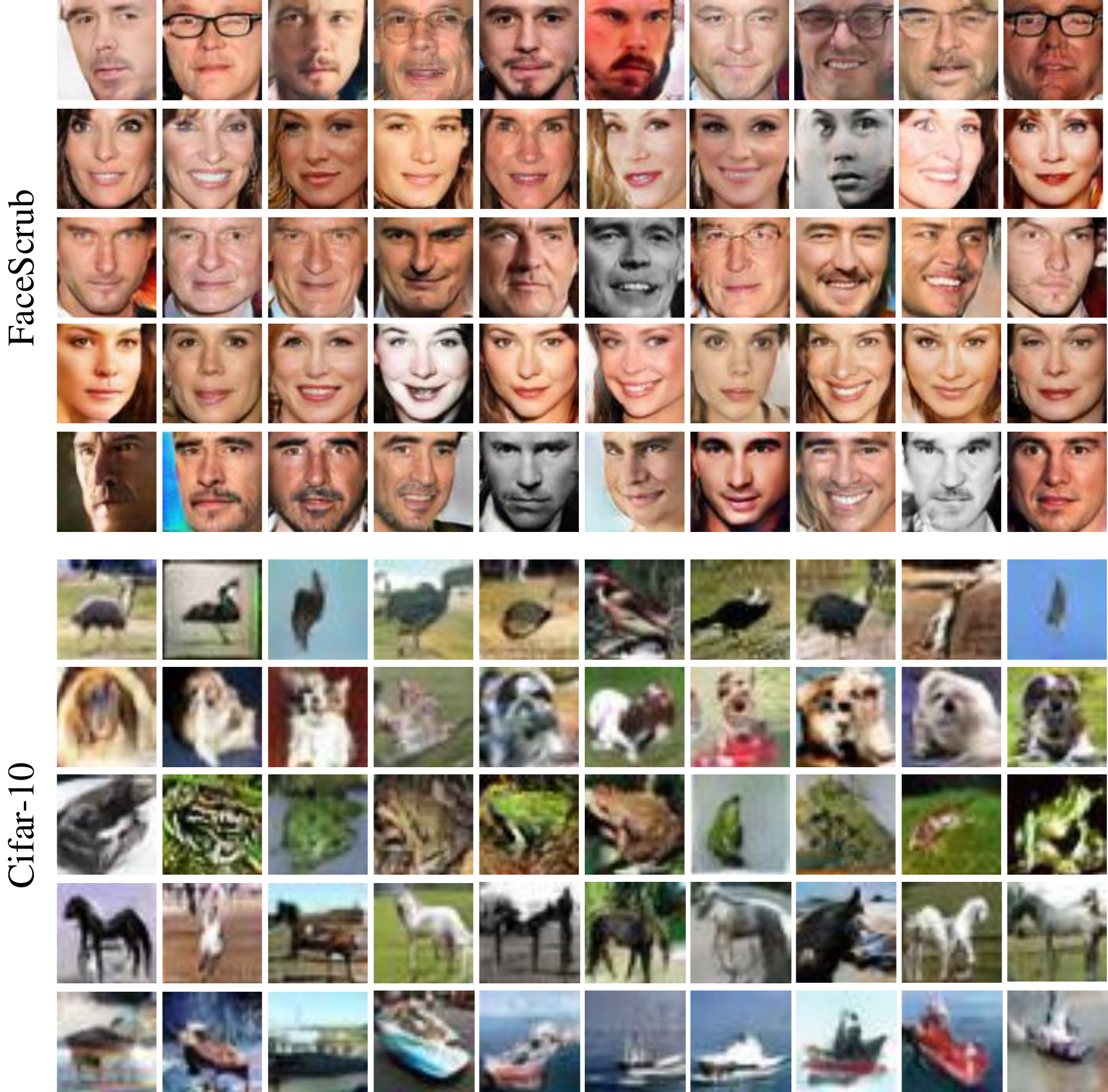}
\end{center}
   \caption{Generated samples of our method on FaceScrub and Cifar-10. Each row shows images of a certain class.}
\label{fig:generated_face_cifar}
\end{figure*}

Figure~\ref{fig:generated_face_cifar} shows generated samples of our method on FaceScrub and Cifar-10 with each row corresponding to a certain class.

\subsection{Additional experiments on CUB-200-2011 and Cifar-100}
We additionally apply our method to CUB-200-2011~\cite{WahCUB_200_2011} and Cifar-100~\cite{krizhevsky2009learning} dataset. The CUB-200-2011 contains 200 categories of birds with 11,788 images in total. For CUB-200-2011, we crop the images according to the bounding boxes provided by the dataset and resize the cropped images to 64 $\times$ 64. The network structure is just same as it used in FaceScrub. For Cifar-100, we use the same network as in Cifar-10. Generated images are shown in Figure~\ref{fig:generated}. Results of $Inception\ Score$ and intra-class diversity are listed in Table~\ref{tab:3} and Table~\ref{tab:4}, respectively.
\begin{figure*}
\begin{center}
\includegraphics[width=1\linewidth]{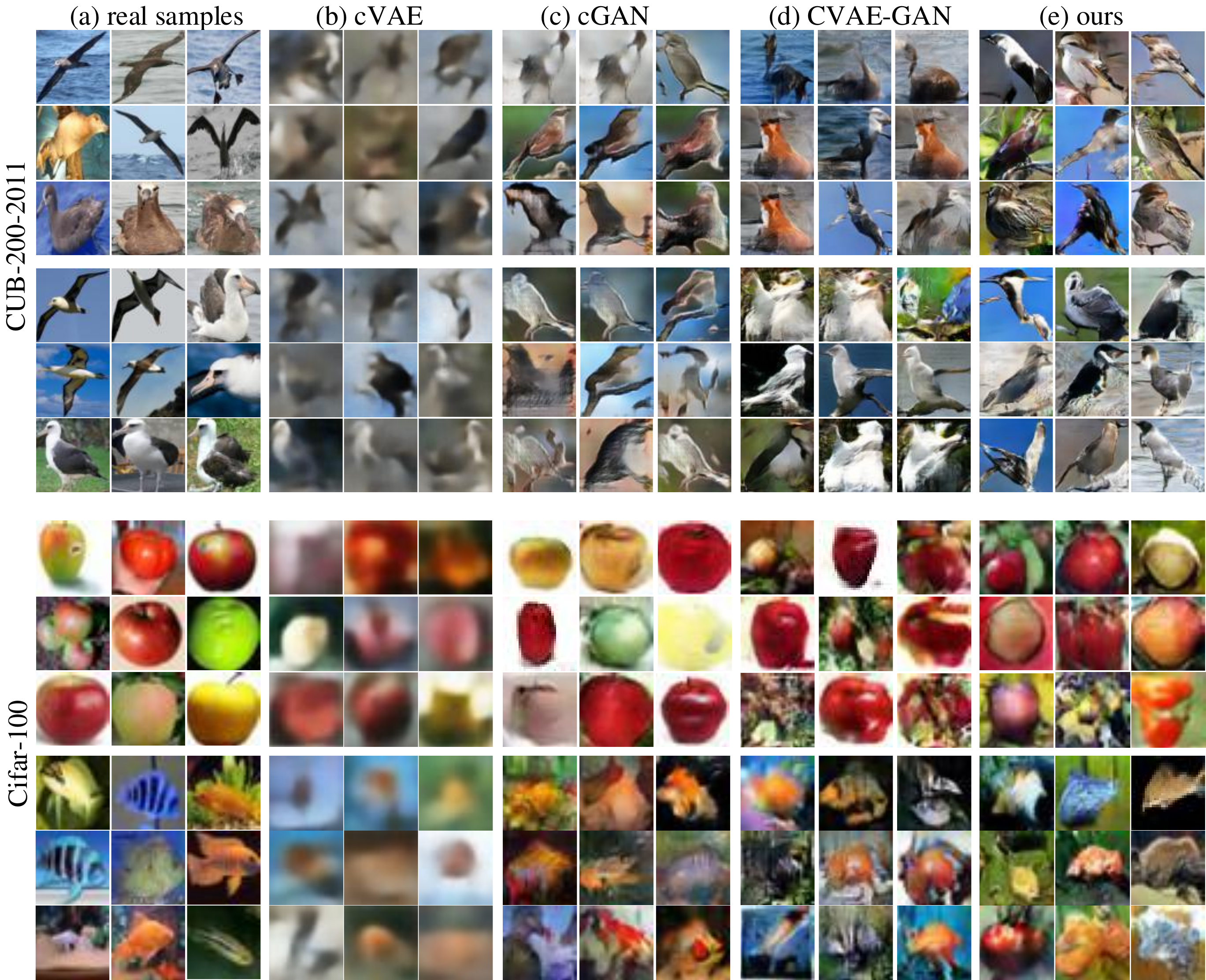}
\end{center}
   \caption{Visualization of generated images of different models on CUB-200-2011 and Cifar-100.}
\label{fig:generated}
\end{figure*}
\begin{table}[htb]
\begin{center}
\begin{tabular}{|l|c|c|}
\hline
 &CUB-200-2011 & Cifar-100\\
\hline\hline
 cVAE~\cite{sohn2015learning}& 37.34& 3.10 \\
 cGAN~\cite{miyato2018cgans}& 78.06 & 6.39\\
 cVAE-GAN~\cite{bao2017cvae}& 91.14 & 6.68\\
 ours& \bf{100.86} & \bf{6.70}\\
\hline
\end{tabular}
\end{center}
\caption{$Inception \ Score$ of different methods.}
\label{tab:3}
\end{table}

\begin{table}[htb]
\begin{center}
\begin{tabular}{|l|c|c|}
\hline
 &CUB-200-2011 & Cifar-100\\
\hline\hline
 cVAE-GAN~\cite{bao2017cvae}& \bf{0.0195}& 0.0179 \\
 ours& 0.0192&\bf{0.0190}\\
\hline
\end{tabular}
\end{center}
\caption{Intra-class diversity of different methods.}
\label{tab:4}
\end{table}
\end{appendices}
\end{document}